\documentclass[runningheads]{llncs}

\usepackage{eccv}

\usepackage{subcaption}

\usepackage{eccvabbrv}

\usepackage{graphicx}
\usepackage{booktabs}

\usepackage[accsupp]{axessibility}  

\usepackage[linesnumbered,ruled,vlined]{algorithm2e}
\usepackage{siunitx}

\usepackage{hyperref}

\usepackage{orcidlink}

\usepackage{wrapfig}

\usepackage{xspace}
\newcommand{\bench}{KilometerVision\xspace}

\usepackage{xcolor}

\usepackage{marvosym}

\makeatletter
\newcommand{\printfnsymbol}[1]{%
  \textsuperscript{\@fnsymbol{#1}}%
}
\makeatother

\begin{document}

\title{\bench: A New Frontier for Large-Scale Spatial Intelligence in VLMs} 

\titlerunning{KilometerVision}

\author{Aravindh Mahendran\inst{1}\thanks{Core contributors} \and 
Michael King\inst{2}\printfnsymbol{1} \and 
Matthew Koichi Grimes\inst{2}\printfnsymbol{1} \and 
Antoine Yang\inst{2} \and 
Tyler Zhu\inst{2,4}\thanks{Work done during internship at Google DeepMind.} \and 
Joseph Heyward\inst{2} \and 
Tengda Han\inst{2} \and 
Shiry Ginosar\inst{5} \and 
Chen Sun\inst{3} \and 
Dima Damen\inst{2} \and 
Simon Osindero\inst{2} \and 
Noah Snavely\inst{3} \and 
Simon Lynen\inst{6} \and 
Jo\~ao Carreira\inst{2} \and 
Viorica P\u atr\u aucean\inst{2}\textsuperscript{(\Letter)}}

\authorrunning{A.~Mahendran, M.~King, M.K.~Grimes et al.}

\institute{Google DeepMind, Berlin, Germany\\
\email{aravindhm@google.com}
\and
Google DeepMind, London, UK\\
\email{\{mjking, mkg, antoineyang, heywardj, tengda, ddamen, osindero, joaoluis, viorica\}@google.com}
\and
Google DeepMind, USA\\
\email{\{chensun, snavely\}@google.com}
\and
Princeton University, Princeton, USA\\
\email{tylerzhu@cs.princeton.edu}
\and
Toyota Technological Institute at Chicago, Chicago, USA\\
\email{shiry@ttic.edu}
\and
Google, Zurich, Switzerland\\
\email{slynen@google.com}}

\maketitle

\begin{figure}[t] 
    \centering
    \begin{minipage}{0.46\textwidth}
        \centering
        \includegraphics[width=\textwidth]{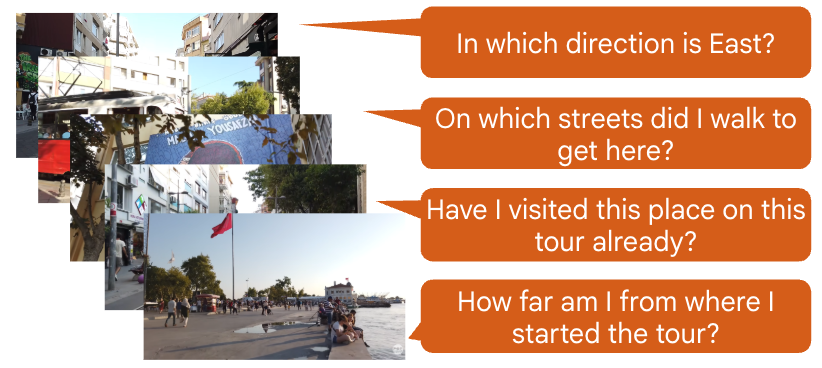}
    \end{minipage}%
    \hfill
    \begin{minipage}{0.25\textwidth}
        \centering
        \includegraphics[width=\textwidth]{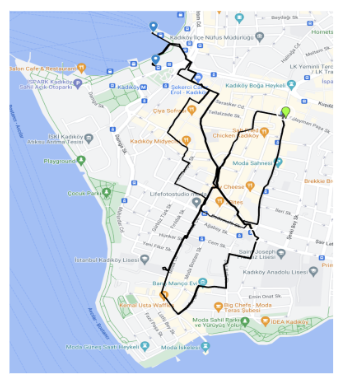}
    \end{minipage}%
    \hfill
    \begin{minipage}{0.25\textwidth}
        \centering
        \includegraphics[width=\textwidth]{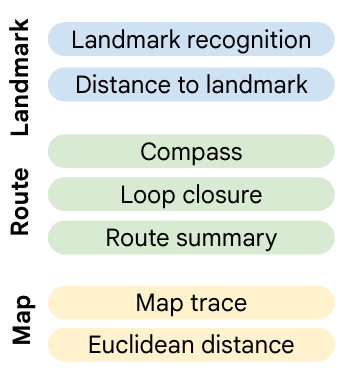}
    \end{minipage}
    \caption{\textbf{\bench} -- the first benchmark probing city-scale spatial understanding from real-world videos spanning 1km distances. \textit{Left:} Real-world hour-long video of a walking tour in Istanbul and relevant questions one would ask when visiting an unfamiliar place. \textit{Centre:} Grounding the video on the map instantly reveals loop closures, distances between landmarks, or sequences of streets visited. \textit{Right:} Using the \textsc{landmark-route-map} paradigm from cognitive science, we define tasks to comprehensively evaluate visual city-scale spatial intelligence in video-language models.}
    \label{fig:teaser}
\end{figure}

\begin{abstract}
  We push the frontier of large-scale spatial intelligence in Vision-Language Models (VLMs) and introduce the first benchmark that probes geographical layout understanding from real-world videos, spanning up to 1km distances. Inspired by the cognitive science literature, we evaluate models against the hierarchical stages of human spatial awareness: anchoring via landmarks, connecting them through routes, and integrating these into global mental maps. Extensive experiments reveal a fundamental divergence in how current AI models process spatial information. Instead of utilising true path integration or forming geometric survey knowledge, we find that VLMs rely almost entirely on 2D visual recognition and text-matching to bypass complex spatial reasoning. 
  
  The benchmark is publicly available at \url{https://perception-test-challenge.github.io/kilometervision.html}.
\end{abstract}

\section{Introduction}
\label{sec:intro}

Spatial intelligence is a hallmark of general intelligence~\cite{multipleintelligences}. Species across the animal kingdom develop tailored spatial representations, intrinsically linked to their dominant sensory modalities, enabling them to recognise places, plan paths, and navigate, sometimes across thousands of kilometres~\cite{hyppo,naturecompass,animalnavigation}. For example, birds rely on geomagnetic fields for orientation, insect-eating bats use echolocation to extract depth, and humans leverage visual and proprioception senses to form mental maps of their environments~\cite{SIEGEL19759,simmons1973,taxonomywayfinding}. Given that current large vision-language models (VLMs) are primarily trained on passive internet-scale data, how do they represent their surroundings?

Multiple works have addressed this question in small-scale environments, studying how VLMs represent objects and their relations~\cite{ramakrishnan2025does,patraucean2023perception}. Navigation and planning capabilities have also been studied in simulated interactive environments or factory-like settings~\cite{song2024towards,yang2024think,li2025unfoldingspatialcognitionevaluating,ramakrishnan2025does}. In this work, we push the envelope to city scale and explore if VLMs can infer valid geographic spatial representations from real-world videos of hour-long walking tours.

We present \textit{\bench}, the first benchmark to comprehensively assess VLMs' city-scale spatial intelligence from real-world videos. We take inspiration from the \textsc{landmarks-route-map} paradigm from the large-scale spatial cognitive literature~\cite{SIEGEL19759,activapassivenavigation,Kim2021Acquisition} and define tasks in multiple-choice video QA format to comprehensively evaluate the existence of these three types of spatial constructs: \textit{landmarks} (ability to identify and localise salient objects), \textit{routes} (egocentric turn-by-turn representations linking two locations), and \textit{survey knowledge} (allocentric maps integrating multiple routes)~\cite{Tolman1948CognitiveMI}; see Fig.~\ref{fig:teaser} (right).

\bench contains 1000 5-way video QAs, defined over 235 YouTube videos. For each question, the model receives a (segment of a) walking tour video together with a question about the video and 5 possible options, out of which only one is correct. The questions are carefully formulated to probe VLMs' visual spatial capabilities rather than their semantic knowledge about various places. 
We evaluate five state-of-the-art VLMs: Gemini 2.5 Flash~\cite{comanici2025gemini25pushingfrontier}, PLM-8B~\cite{cho2025perceptionlmopenaccessdatamodels}, Qwen2.5-VL-72B~\cite{bai2025qwen25vltechnicalreport}, Claude Opus~\cite{anthropic2024claude3}, GPT-5~\cite{OpenAI_GPT5}. While our experimental results show a large performance variation across model families and model sizes, a consistent limitation emerges across all models: rather than utilising true path integration or building geometric mental maps, current VLMs predominantly bypass complex spatial reasoning by relying on 2D visual recognition and semantic text-matching.

\section{Related work}
\label{sec:relatedwork}

\noindent{\textbf{Large multimodal models and evaluation:}} Recent advances in large multimodal models (VLMs)~\cite{ko2023large, zhang2023video,2023videochat, li2024llava, damonlpsg2024videollama2,comanici2025gemini25pushingfrontier} are often achieved by integrating high-dimensional, continuous sensory signals (\eg, visual and audio) into compact representations as inputs to large language models. In order to capture visual information, some approaches~\cite{zhang2023video, damonlpsg2024videollama2, ko2023large, xue2024xgen, li2024llava} employ pre-trained image encoders~\cite{zhai2023sigmoid, oquab2024dinov} to extract representations from individual frames, whereas other approaches~\cite{wang2022internvideo, Maaz2023VideoChatGPT} utilize video encoders to extract (short-term) spatiotemporal representations. Beyond distributed representations, it has been demonstrated~\cite{2023videochat,wang2023vamos,wang2024lifelongmemory} that interpretable video representations, such as dense captions, often capture sufficient information for many existing video question answering benchmarks, such as NextQA~\cite{xiao2021next} and EgoSchema~\cite{EgoSchema}.

The success of VLMs has inspired researchers to introduce more challenging benchmarks beyond action classification~\cite{carreira2017quo} and localization~\cite{caba2015activitynet,gu2018ava}. One such attempt aims to evaluate multimodal concept abstraction and spatiotemporal reasoning, in either synthetic~\cite{girdhar2020cater,Yi2020CLEVRER}, egocentric~\cite{Damen2021TheED,Grauman2021Ego4DAT}, or carefully curated real-world scenarios~\cite{patraucean2023perception}. Another attempt focuses on long-form video understanding, where the videos are usually sourced from movies~\cite{tapaswi2016movieqa,rawal2024cinepile}, or procedural demonstrations~\cite{zhukov2019cross,EgoSchema,Grauman2021Ego4DAT} and vlogs~\cite{fouhey2018lifestyle}. Finally, a trend in recent benchmarks aims at offering a comprehensive evaluation on various aspects of multimodal perception and reasoning from diverse domains, such as VideoMME~\cite{fu2025video} and LVBench~\cite{wang2025lvbench}.

\medskip
\noindent{\textbf{Benchmarking spatial intelligence:}}
Traditionally, spatial understanding was evaluated through low-level tasks like depth estimation~\cite{Geiger2012CVPR,Silberman:ECCV12}, pose estimation~\cite{poseest}, or 3D reconstruction~\cite{6907054,6630703,Krishnan2025ICCV}, tackled by specialised methods (SFM~\cite{sfm}, SLAM~\cite{6907054,6630703}). Given VLMs' potential to become general perception models, more and more efforts are focusing on evaluating their spatial capabilities using a language interface, often taking inspiration from the cognitive literature to design challenging tasks~\cite{patraucean2023perception,yang2024think,li2025unfoldingspatialcognitionevaluating,ramakrishnan2025does}. The main question that these works try to answer is if the current internet-scale training datasets and strategies can allow spatial intelligence to emerge~\cite{vafa2024evaluating,ramakrishnan2025does,yang2024think,song2024towards,li2025unfoldingspatialcognitionevaluating,planbench}, can VLMs learn a valid world model of physical environments by learning from passively observed data.

\medskip
\noindent{\textbf{Large-scale geographic capabilities:}} Several works have tried to push the envelope of spatial understanding to geographic scale, due to the unique challenges that such settings expose in terms of memory and long-context understanding. The authors of~\cite{vafa2024evaluating} study the world model learnt by an LLM from a large-scale text taxi rides dataset. Still using only text modality, the authors of~\cite{Roberts2023Gpt4geo,tooluse} design tasks probing geographic knowledge in LLMs. In~\cite{charting} images are provided as inputs as well. Other works are attempting to place VLM agents in interactive environments to probe their planning and navigation capabilities~\cite{touchdown,DBLP:journals/corr/abs-2502-11163,xu2025flame}. Our proposed benchmark is novel and complements these existing benchmarks, being the first to use real-world videos of walking tours~\cite{venkataramanan2023imagenet} to comprehensively probe VLMs' capabilities to recognise landmarks, routes, and build mental maps of cities from videos alone.

\section{Video dataset}
\label{sec:benchmark}

\subsection{Video source}
We rely on high-resolution real-world hour-long YouTube Walking Tour videos filmed in various cities around the world. In these videos, the camera wearer walks around the city, visiting landmarks, sometimes returning to places already visited (\ie~loop closures). We start from the nine city-life videos in the Walking Tours dataset~\cite{venkataramanan2023imagenet}, and select 226 extra YouTube videos with similar characteristics (high-resolution, hour-long), prioritising diverse coverage of cities in different countries, for a total of 235 videos with about 288 hours of video data. 

\subsection{Grounding videos on the map}
\label{sec:vps}
To facilitate extracting annotations for our tasks, we first design an efficient pipeline to ground the videos on the map, \ie extract the (\textit{latitude}, \textit{longitude}) coordinates and \textit{camera pose} of each frame in the video at a given frame rate, \eg 1 FPS. This produces detailed accurate route traces on the map (see Fig.~\ref{fig:teaser}, centre), which instantly reveal important aspects that would otherwise require watching a long video, possibly multiple times, to discover, \eg loop closures when walking around in a neighbourhood and returning to the same place on a different path. Given such map traces, we can then use simple heuristics and minimal human annotations to extract at scale ground-truth annotations for various spatial tasks. 

We leverage Google's \textit{Visual Positioning System} (VPS)~\cite{Google:Patent:VPS}, a public API that relies on StreetView imagery to estimate the location and camera pose of any query image. Given a query image, VPS calculates its embedding using an image encoder, then retrieves the (lat, lon) coordinates and camera pose of the nearest neighbour image stored in Google StreetView embedding. For this operation to be efficient, an estimate of the location is also necessary (\eg within 100m of the true location). Given this constraint and to run efficiently on long videos,  we rely on human raters to provide the (lat, lon) coordinates corresponding to the start and end of each walking tour video in our dataset. Then, we run the VPS API on consecutive video frames at 1FPS, updating the location estimate with the newly returned location as the video progresses.

In more detail, for each given frame, the VPS output is a tuple $(\text{lat}, \allowbreak \text{lon}, \allowbreak \text{rot}, \allowbreak \text{trans}, \allowbreak \text{confidence})$, where (latitude, longitude) are real scalar values, $\text{rot}$ represents the camera rotation as a quaternion in the Earth-Centered, Earth-Fixed (ECEF) coordinate system $\text{rot}\in\mathbb{R}^{4}$, $\text{trans}$ is the camera translation as a matrix $\mathbb{R}^{1\times3}$, and the confidence score is between 0.0 and 1.0. If this confidence score is larger than a given threshold, we update the estimated location with the newly returned location, otherwise we keep the existing estimate when processing the next frame. To improve the quality of the VPS predictions, we repeat the operation but in reverse order using the last frame’s human-annotated position as initial estimate and making VPS calls consuming the video frames in reverse. The two passes are then merged by using a confidence-weighted average between the forward pass predictions and the reverse pass. Finally, a last VPS run is executed, using each frame's weighted average as initial estimate for that frame. Note that, despite the repeated runs, this is a robust pipeline that can be run at scale on thousands of hour-long videos; for our 288 hours of videos, it takes about a day to do the full processing. To circumvent VPS errors in feature-poor areas~\cite{horvath2025investigating}, we apply outlier removal as detailed in the appendix.

To validate this approach of grounding videos on the map, we compared the obtained traces against traces collected by humans using Dynamic Time Warping (DTW) as a metric. To calibrate the DTW metric, \ie get a sense of what is an acceptable error between human-collected ground truth and an automatically extracted VPS path, we collected 2 parallel human annotations for a set of 10 hour-long videos ($\sim$15 hours of video data) using Google Maps as annotation interface and asking human raters to draw on the map the path followed by the camera wearer. Twelve more videos ($\sim$14 hours of data) were used for validation. Overall, about 10\% of the data has human annotations.
For each video, we calculated the DTW distance between the two human-provided paths, giving us a measure of acceptable variability between two paths representing the same walking tour. We then calculated DTW distances between VPS traces and human-provided traces, and discarded segments of the VPS trace where the DTW metric was above the threshold; more details are included in the appendix. It is worth noting that the cost reduction offered by our pipeline is massive. Human raters need only $\sim$5 minutes per video to mark the location of the start and end of the video, then the video is grounded automatically. Without our pipeline, drawing manually the full path on the map took $\sim$6 hours per 1h-long video and it only provides position information, not pose.

In a recent parallel work~\cite{Krishnan2025ICCV}, the authors relied on ARIA glasses to similarly ground walking tour videos on the map and design a city-scale benchmark for SLAM systems. Compared to this work, which requires a person to wear the ARIA glasses and walk around to collect the videos, our system allows us to ground at scale any \textit{already-filmed} video, with any camera, anywhere in the world.

\section{Benchmark tasks}

As evidenced in the cognitive literature, continued exposure to a new environment leads to three stages of spatial awareness in humans: \textsc{landmark-route-map}~\cite{activapassivenavigation}. At the landmark stage, we observe salient objects (\eg a clock tower or a statue) that we can easily recognise and allow us to anchor ourselves in space. At the route stage, we infer sequences of turns that allow us to move from one landmark to another. Finally, the mental map stage corresponds to an overall layout of the environment that allows us to derive knowledge not readily observed while interacting with the environment, \eg shortcuts. 
We define evaluation tasks at these three levels of representation to probe if VLMs develop similar constructs. We rely on multiple-choice video QA format because modern VLMs process and generate text natively, providing a standardised and accessible interface for evaluation. Importantly, this format grants us access to the models' intermediate ``thinking traces'', allowing us to qualitatively investigate the actual strategies they use to solve spatial problems. In contrast, attempting to explicitly extract continuous 3D spatial representations like metric depth maps, 3D point clouds, or dense topological graphs, directly from the latent spaces of general-purpose VLMs, would be highly non-trivial and model specific.

\begin{figure}[t]
    \centering
\includegraphics[width=0.3\linewidth]{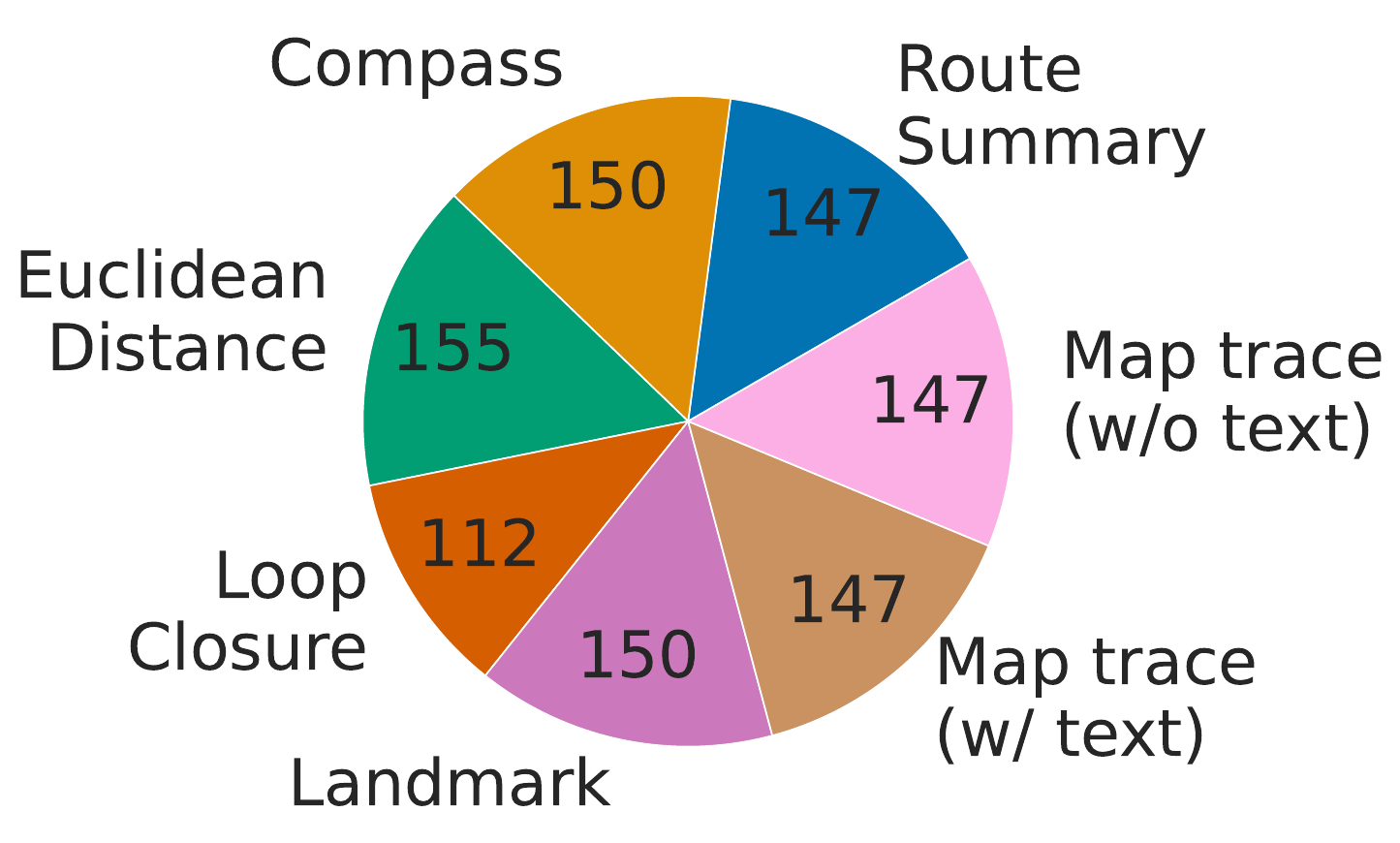}~%
\includegraphics[width=0.25\linewidth]{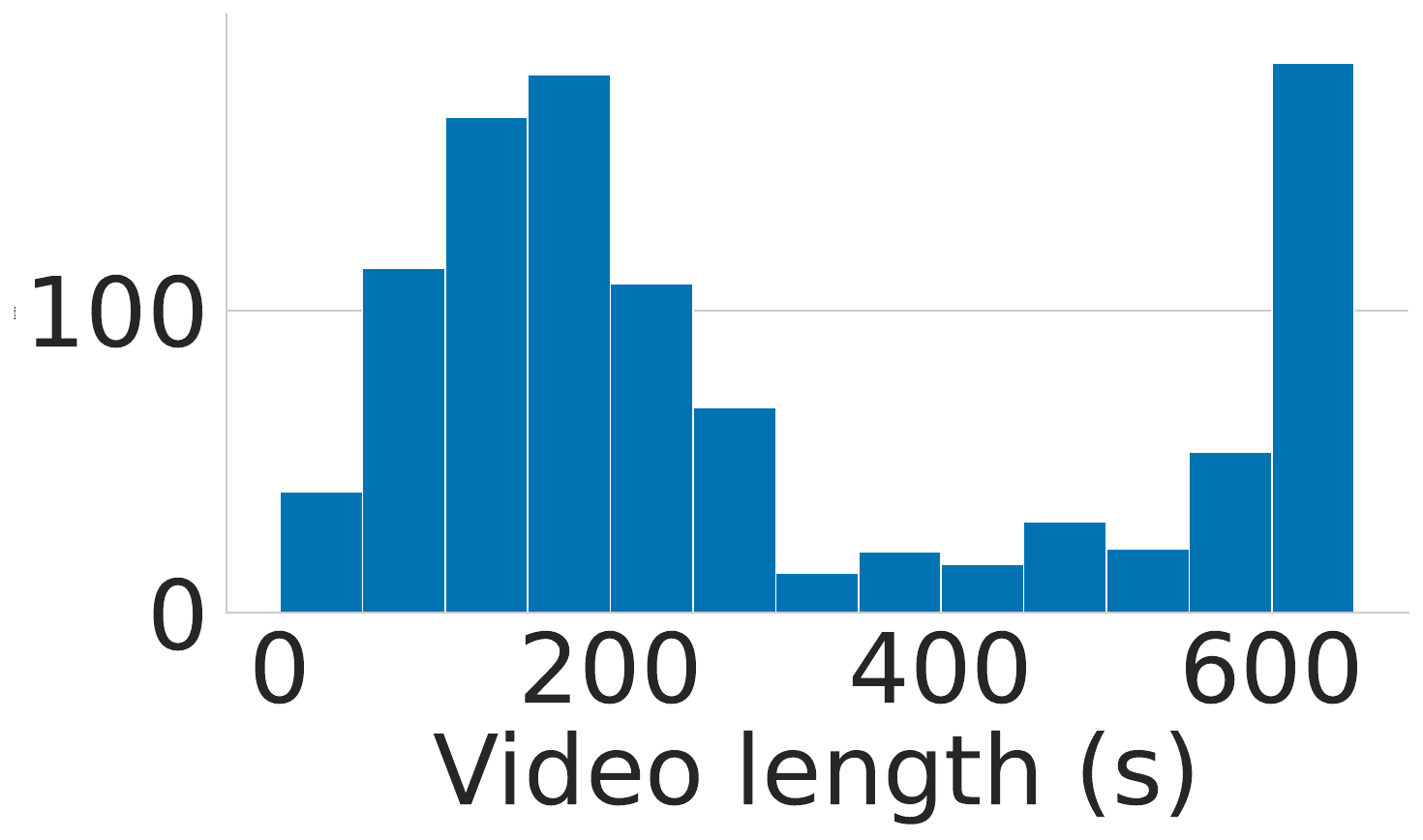}
    \caption{Left: Distribution of question types in the \bench benchmark. Right: Histogram over video lengths across question types.}
    \label{fig:composition}
\end{figure}

To keep the evaluation efficient, we define the tasks on video segments of up to 10 minutes long. At average walking speed, a 10-minute segment corresponds to about 1km distance traversed; see examples of corresponding map traces in Fig.~\ref{fig:loops}. 
We include up to 150 QAs for each type of task, again for efficiency reasons. But note that given our pipeline to ground videos on the map, we can efficiently generate a larger number of questions for each task (with the exception of loop closure, as loops occur less frequently in walking tours). Fig.~\ref{fig:composition} shows the distribution of tasks and video lengths in the benchmark. When compared to prior work, \eg VSI-Bench, our videos are 3-6$\times$ longer, covering much larger spatial areas (factory \vs neighbourhood).        

\subsection{Landmarks}
\noindent{\textbf{Landmark recognition:}}
We define landmark recognition tasks by using a non-standard multiple-choice video QA format. The model is presented with a text question, a video clip, and a picture of a landmark that may or may not have appeared on the path shown in the clip. The questions are of the form: 

\noindent{\textit{Does the image show a location in the video, and if so, what is the straight-line distance between the camera location of the image and the camera location of the final video frame?}}

\begin{figure}
  \centering 
  \includegraphics[width=0.42\linewidth]{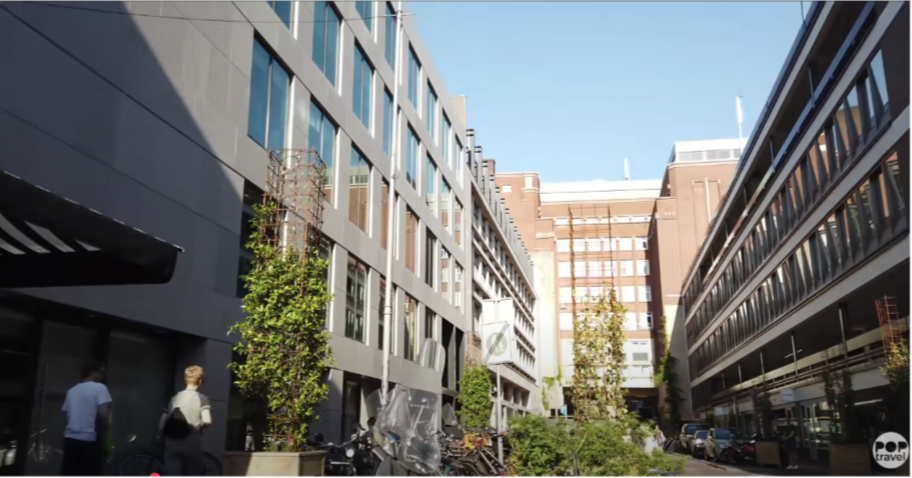}%
  \includegraphics[width=0.42\linewidth]{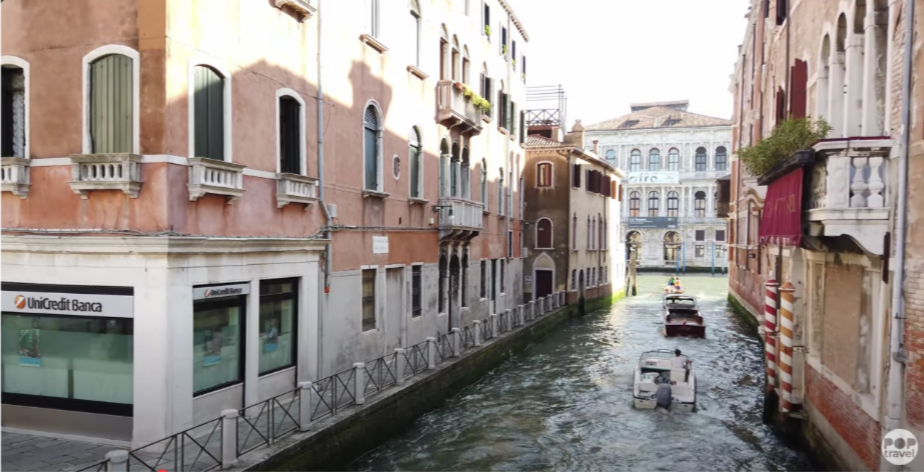}
  \caption{Examples of landmark frames (Left: Amsterdam, right: Venice).}
  \label{fig:landmarks}
\end{figure}

\begin{figure*}
    \centering
    \includegraphics[width=0.245\linewidth]{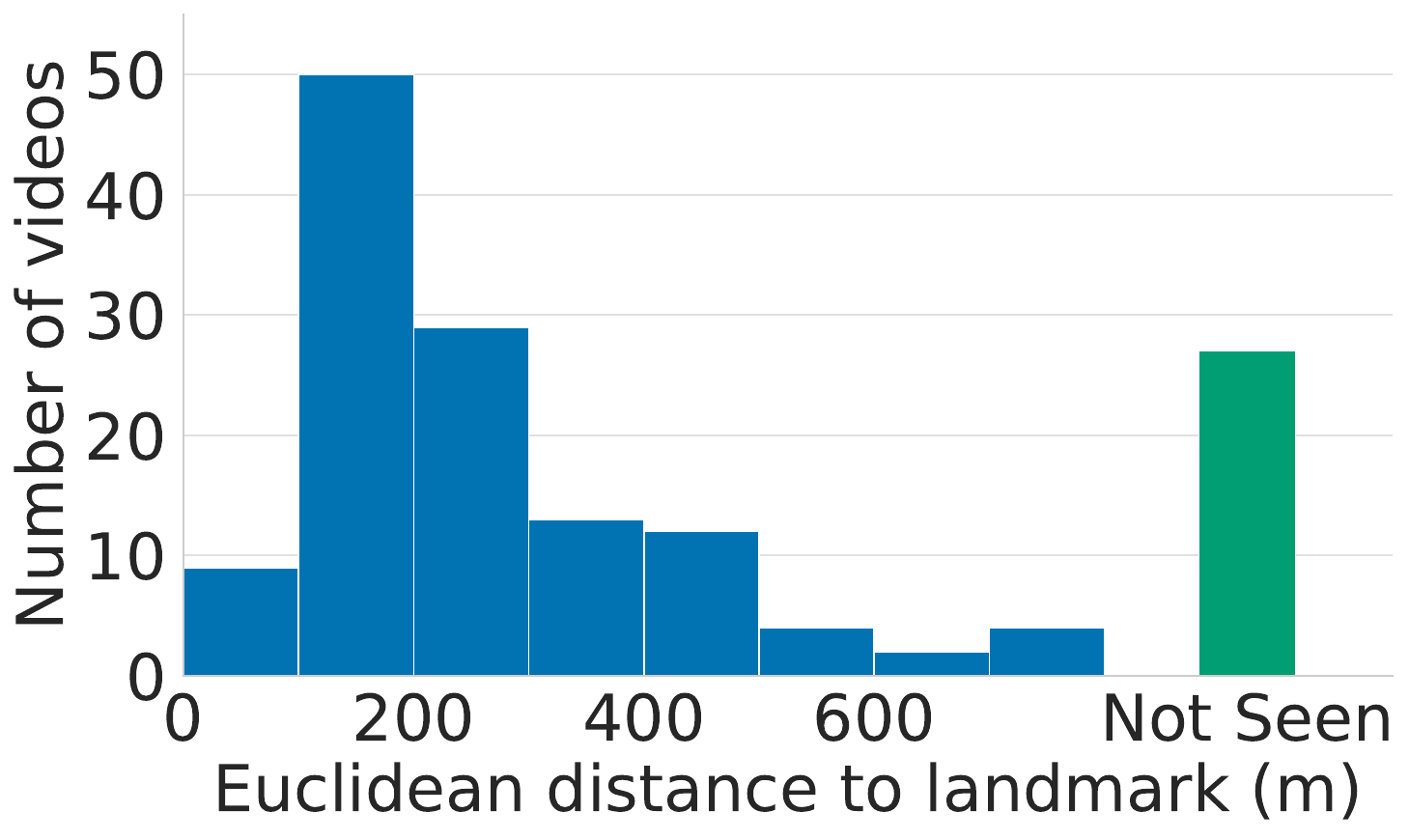}%
    \includegraphics[width=0.245\linewidth]{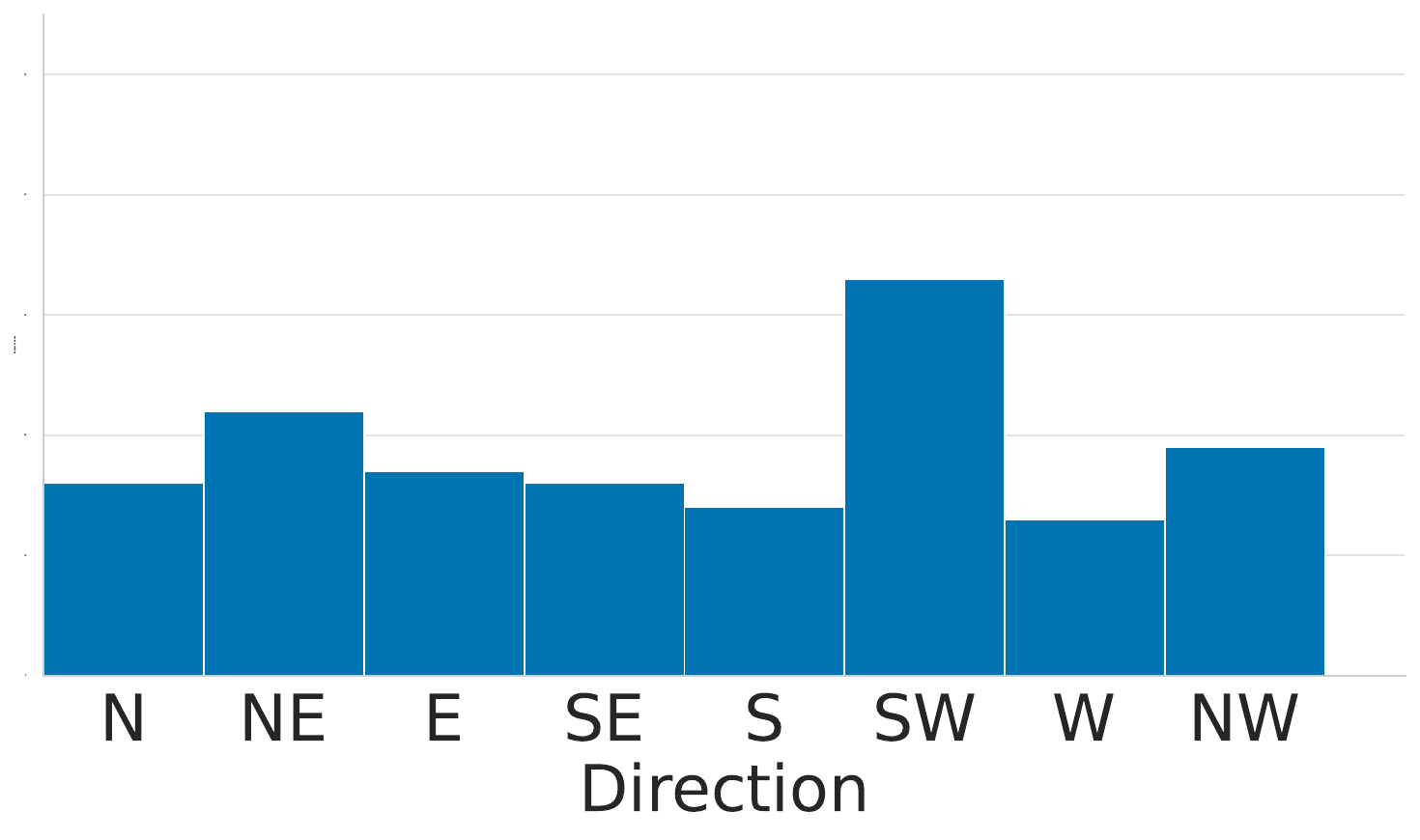}%
    \includegraphics[width=0.245\linewidth]{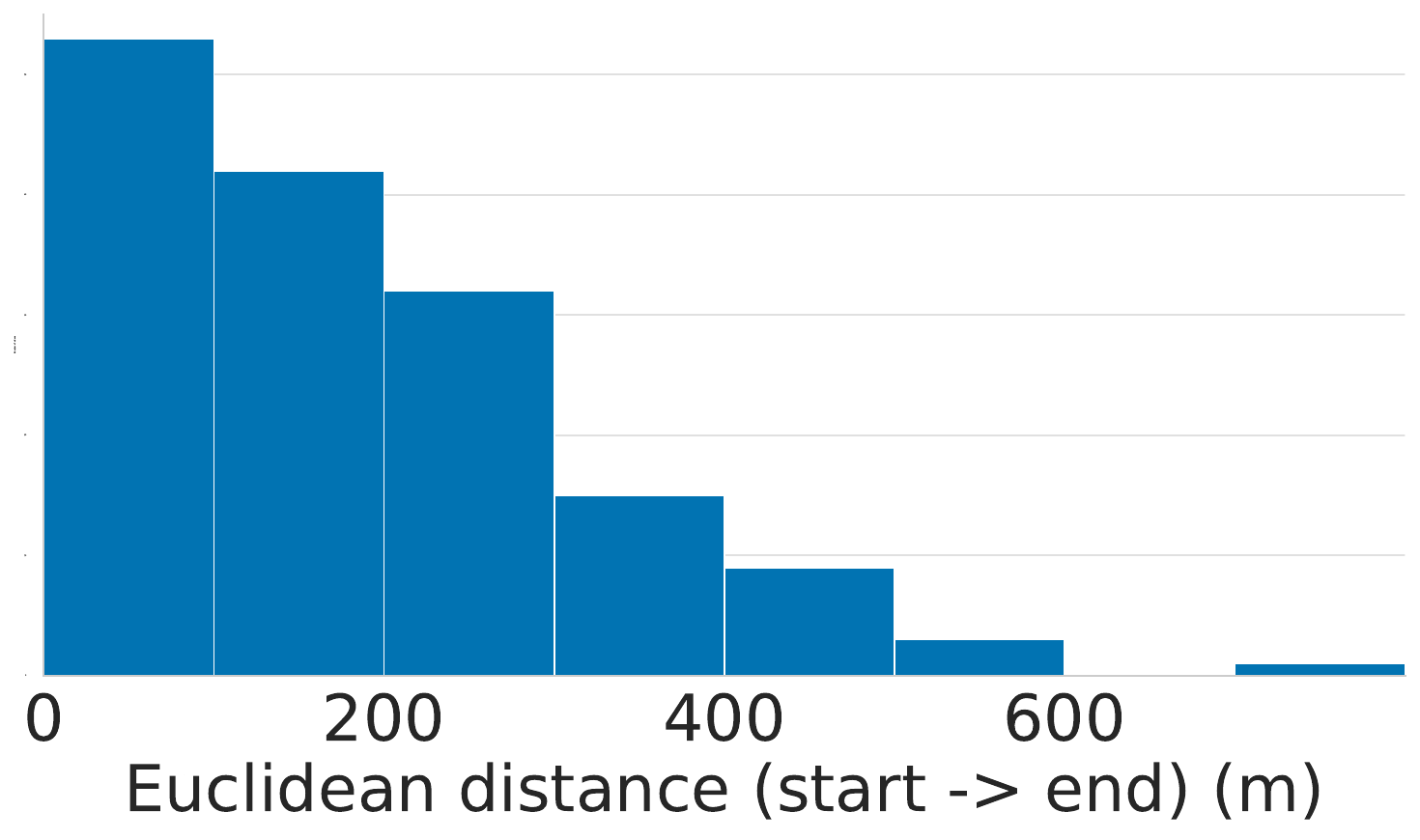}%
    \includegraphics[width=0.245\linewidth]{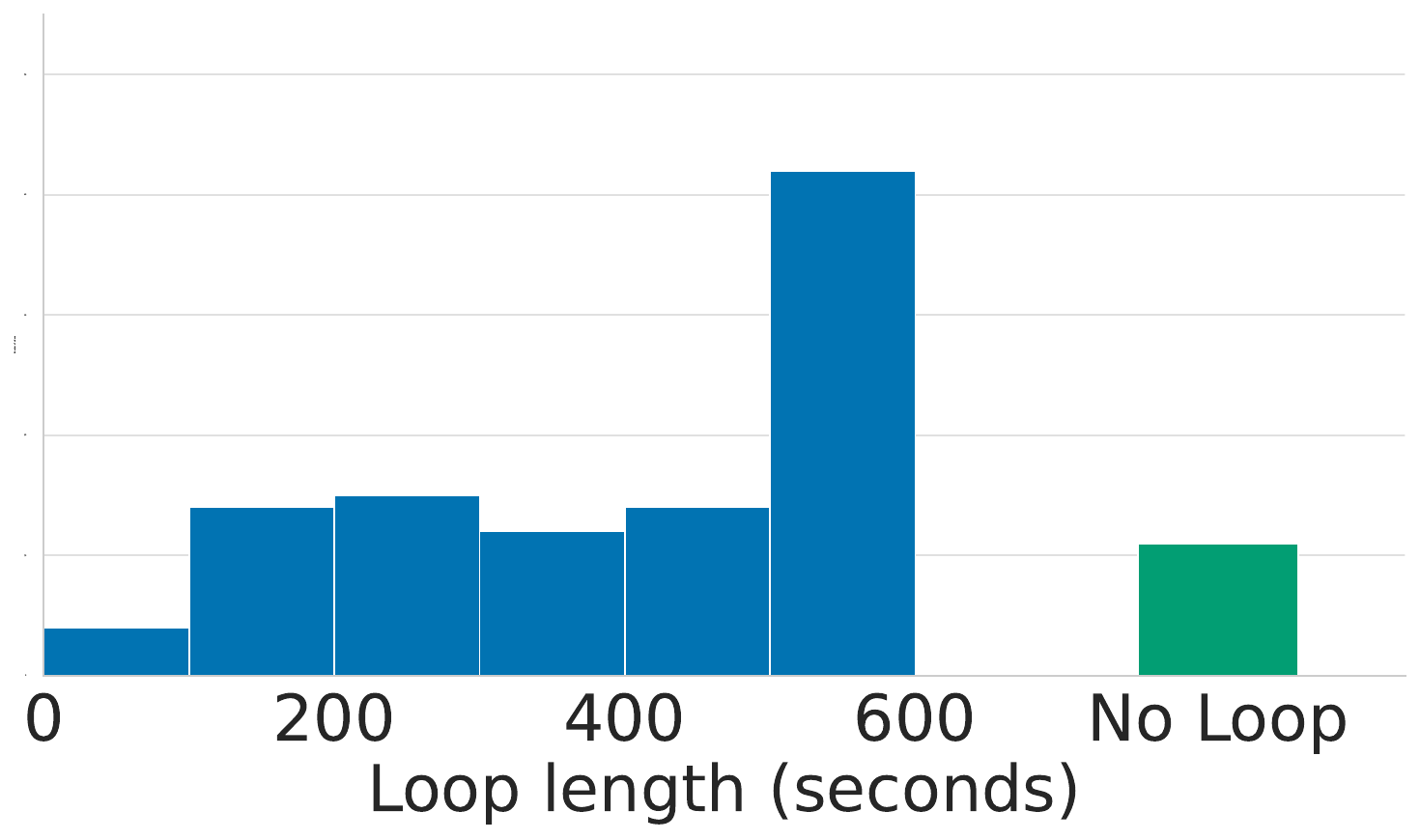}
    \caption{Left: Distribution of line-of-sight distances (in metres) to landmarks used in our dataset. We also include questions with landmarks that haven't been visited on the tour, for which the correct answer is `not seen'. Centre-left: Distribution of orientations at the end of the walking tour for the compass task. Centre-right: Distribution of Euclidean distances between the start and the end of each tour. Right: Distribution of loop lengths in seconds; maximum video length is 600 seconds (10 minutes). We also include 11 question-answer pairs where there is no loop.}
    \label{fig:datasetstats}
\end{figure*}

This is followed by four possible distances in meters (one being the correct one if the landmark is in the given video), plus a fifth \textit{negative} option: \textit{That place was not seen in the video}. For simplicity, we select landmark images from the video frames themselves, by  filtering for video frames that have very high VPS confidence scores ($>.98$), under the intuition that these are the most distinctive; we also apply non-maximal suppression to keep only the peaks and remove duplicates. Examples of landmark pictures obtained after this filtering are included in Fig.~\ref{fig:landmarks}, for Walking Tour videos filmed in Amsterdam and Venice, respectively. Fig.~\ref{fig:datasetstats} shows the distribution of correct answers for landmark distances. We also include questions for which the negative answer is the correct one. In these cases, we extract landmarks from parts of the video that do not overlap with the current segment. Note that this landmark recognition task is an example of needle-in-haystack problem, at which Transformer-like architectures are known to excel~\cite{comanici2025gemini25pushingfrontier}, and our results confirm this (see Sec.~\ref{sec:discussion} and analysis in Fig.~\ref{fig:visual_recognition}, right).

\medskip
\noindent{\textbf{Distance to landmark:}} Note that the second part of the question above related to distance estimation requires map-level knowledge. We use this composite form for efficiency reasons, to run a single model query. In the analysis of the results, we break down the performance into landmark recognition and distance estimation, see Fig.~\ref{fig:visual_recognition}, right.

\subsection{Routes}
\label{sec:routes}

\noindent{\textbf{Compass:}} 
We formulate compass questions related to global camera orientation by giving the orientation of the camera at the beginning of the clip and asking the model to output the orientation at the end of the clip. These allow us to probe the model's ability to perform simplified path integration, necessary for understanding routes and building mental maps of large-scale environments~\cite{1295ae4551b9407e8335d780899875ca}. To correctly answer this question, the model could keep track of point correspondences across frames and estimate the camera motion. Alternatively, the model could infer the cardinal orientation using visual cues, \eg the sun position.
However, such cues are often not available (\eg if the person walks on a narrow street and the sun position cannot be inferred), so the camera motion estimation is a more reliable strategy. We use the following question prompt:  \textit{For this question, you need to keep track of the direction that the camera is facing throughout the given walking tour video. Given that the camera is pointing towards the \texttt{<orientation\_start>} at the start, where is the camera pointing at the very end?}

The orientation is binned into one of eight possibilities: \textsc{North, Northeast, East, Southeast, South, Southwest, West, Northwest}. The ground-truth end orientation and the start orientation are extracted automatically from the VPS predictions after outlier removal. Rotation quaternions in the ECEF reference frame are mapped to the local tangent plane to calculate the yaw with respect to the direction pointing towards north at that location. The 4 incorrect options in the multiple-choice task are sampled at random.  Clips of up to 10-minutes duration are sampled at random such that the start and end frames have high-confidence VPS predictions. We start by extracting a large number of possible questions, and we filter down to keep a subset of 150 QAs by sampling at random questions with distinct start and end orientations; see the distribution of orientations in Fig.~\ref{fig:datasetstats}, centre-left.

\medskip
\noindent{\textbf{Loop closure detection:}} 
This is a critical task in large-scale spatial understanding, commonly used by simultaneous localisation and mapping systems (SLAM)~\cite{slam1,1570189} to correct navigation drift. SLAM solutions rely on a combination of visual recognition and geometric path integration strategies. We formulate loop closure detection questions in multiple-choice format, bridging in this way the gap between the VLM and SLAM communities. 

Given our videos grounded on the map, we extract candidate loops by running a simple heuristic over the VPS predictions, checking if the current VPS point has been visited previously within a 4m radius. Then, we conduct manual inspection of the candidate loops with human raters to remove false positives caused by possible VPS outliers. We use the following question prompt in our benchmark: 

\noindent{\textit{Has the location at the end of the video been visited at an earlier point in the video? If so, when?}}

\begin{figure}[t]
  \centering
  \begin{subfigure}{0.25\linewidth}
    \includegraphics[width=0.98\linewidth]{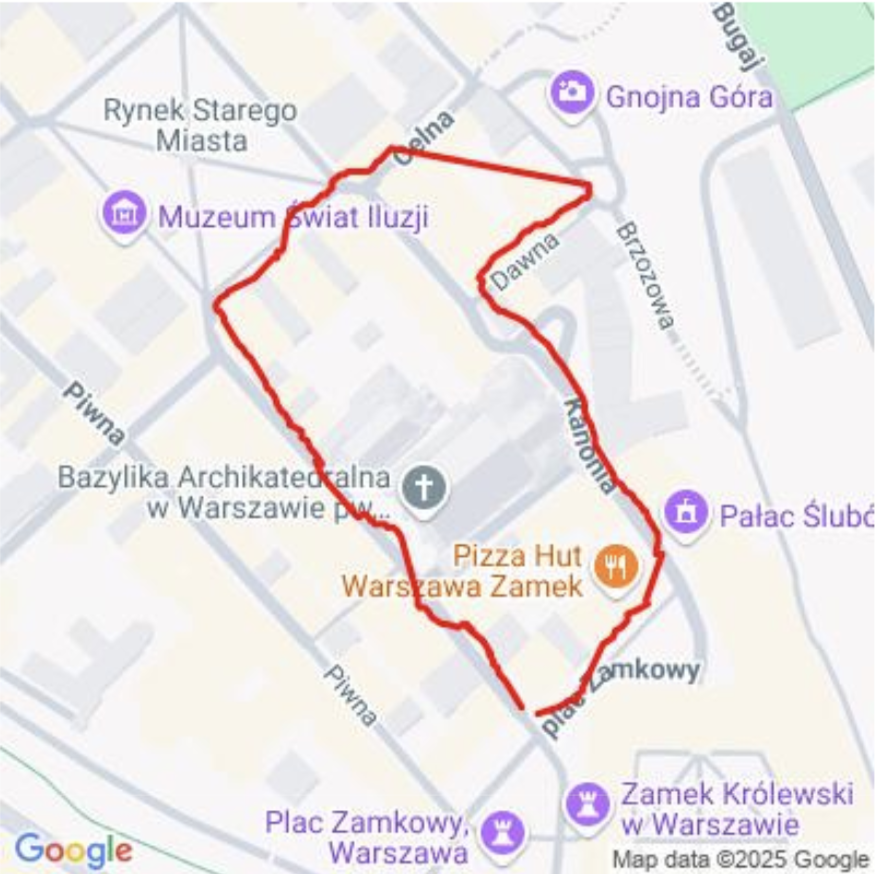}
    \caption{}
    \label{fig:loop-yes}
  \end{subfigure}
  \begin{subfigure}{0.25\linewidth}
    \includegraphics[width=0.98\linewidth]{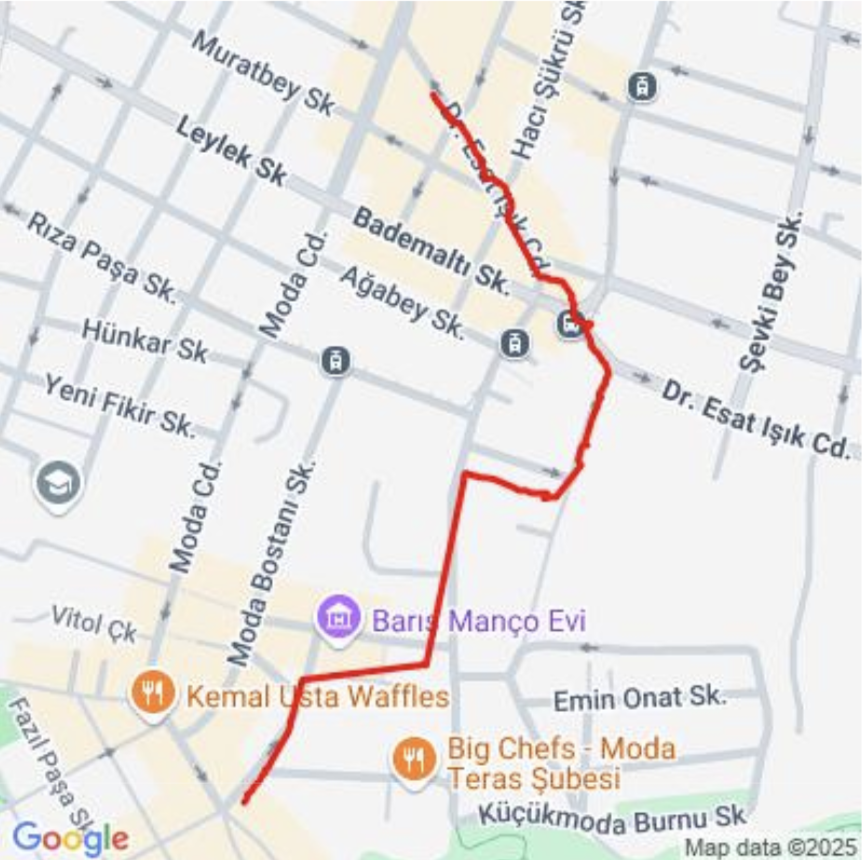}
    \caption{}
    \label{fig:loop-no}
  \end{subfigure}
  \begin{subfigure}{0.25\linewidth}
    \includegraphics[width=0.98\linewidth]{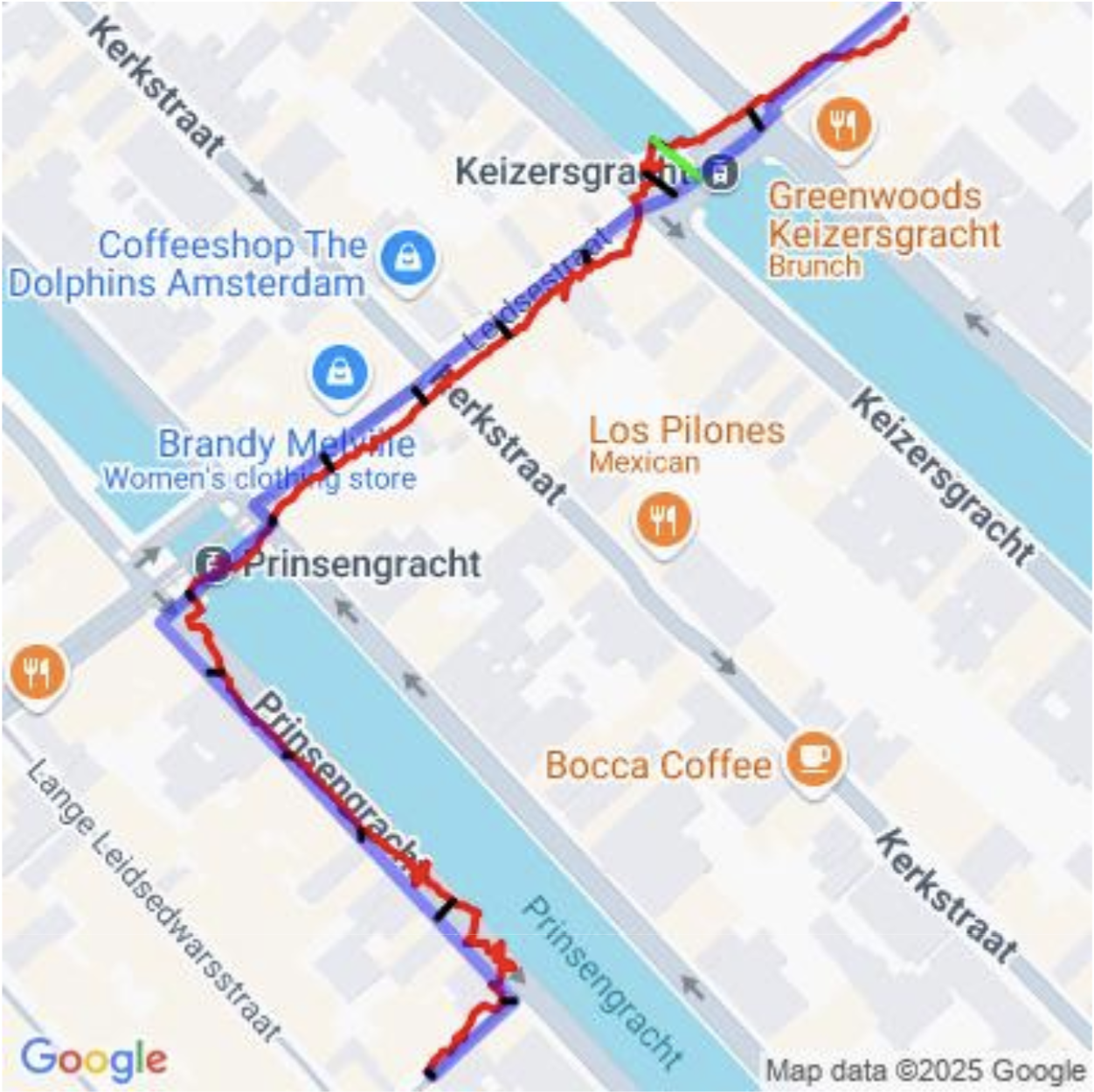}
    \caption{}
  \end{subfigure}
  \caption{(a) Example of a VPS trace with a loop. (b) Example of a VPS trace without a loop. (c) Given the VPS coordinates for a video segment (shown in red), we select high-confidence intermediate VPS points to query the Google Maps Compute Routes API and get a closely-matching trace (shown in blue). Black segments indicate the optimal pairings found by DTW (Dynamic Time Warping) metric. In green, we show the maximum distance (error).}
  \label{fig:loops}
\end{figure}

The negative options are sampled at random making sure that they are at least 1 minute apart from the correct option. Note that we also sample \textit{negative} questions, \ie questions for which the correct answer is \textit{The location has not been visited before in this video.} Figure~\ref{fig:loops} shows trace examples of positive and negative scenarios and Fig.~\ref{fig:datasetstats} shows the distribution of loop lengths in our dataset. We opted to use this composite formulation that includes the time recall to be able to generate five meaningful answer options instead of the binary Yes / No. We break down the performance into loop closure detection and time recall in Fig.~\ref{fig:visual_recognition}, right, for analysis purposes.

\medskip
\noindent{\textbf{Route summary:}}  
We define route summary tasks to probe if the model can identify a summary of the path shown in the video in terms of navigation actions (turn left, turn right, go straight). The question prompt is: \textit{Which of these sets of directions correctly represent the path of the person in this video?}

To generate the ground truth answers, we rely on Google Maps  Routes API that we query using VPS coordinates for start, end, and intermediate points. To ensure that the generated route closely matches the actual path followed in the video (and indicated by VPS), we sample multiple routes using Routes API using different intermediate points and then we use again DTW distance (mentioned in sec~\ref{sec:vps}) to select the best match with the VPS trace; see Fig.~\ref{fig:loops} (c). If the maximum error between optimal pairs found by DTW is not below a given threshold, we discard the question. The pipeline allows us to generate and check a large number of questions very efficiently, so we can afford to select only the questions that satisfy our error constraints. Finally, we keep as correct answer the text description returned by the Routes API for this best match\footnote{We also experimented with fitting a polyline to the VPS predicted locations and deriving the sequence of turns from it, but this led to noisier redundant navigation instructions.}. 

\begin{figure*}[t]
    \centering
    \includegraphics[width=0.95\linewidth]{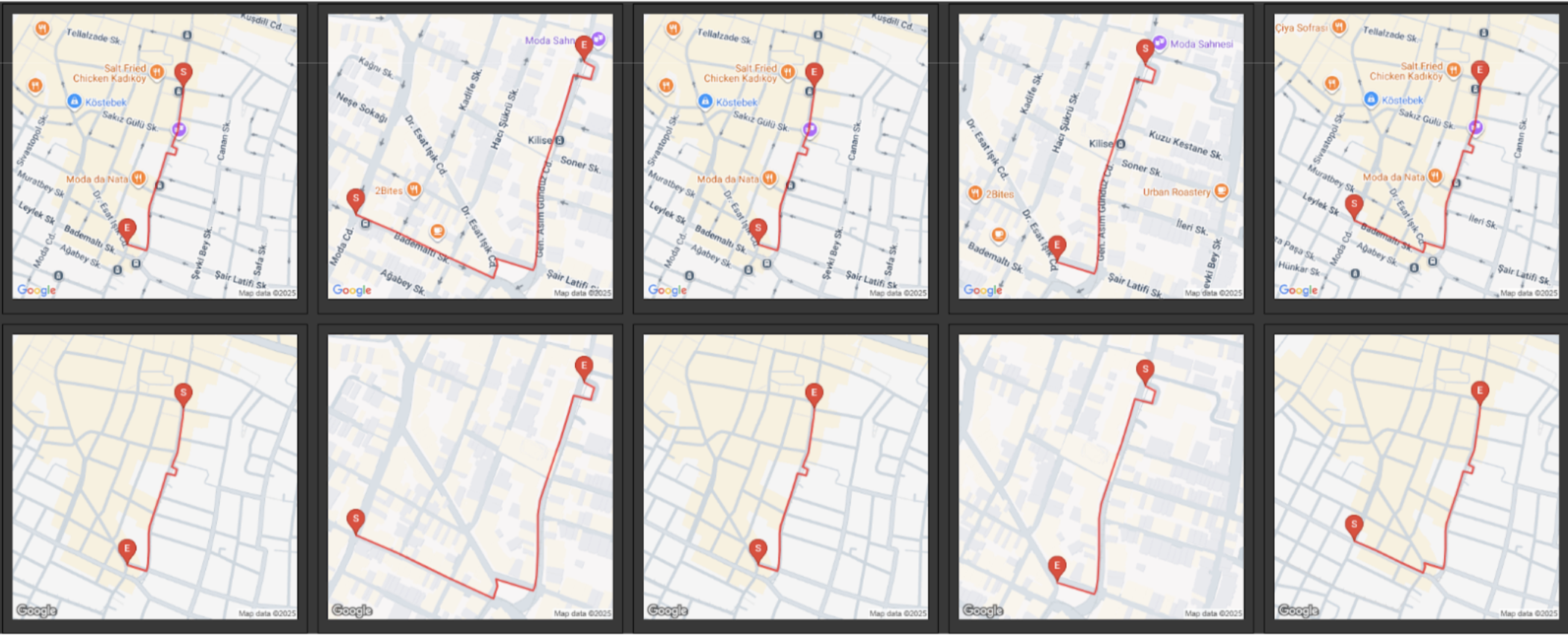}
    \caption{\textbf{Top}: Map trace images used as options for the Map trace task. \textbf{Bottom}: A more challenging map trace task variation where the route traces are in-painted over maps without text labels or landmark icons.}
    \label{fig:maps}
\end{figure*}

To generate the negative options, we consider partially-overlapping segments, forcing the model to correctly identify the start and end point of the tour to discard these. Another negative option that we use is the reversed version of the ground truth path. Fig.~\ref{fig:maps} (top) shows some possible options, which are also used for the map trace task explained below.

As an ablation, we also experimented with an easier version of the task, where we sample negative options from video segments that do not overlap with the ground truth segment at all; see appendix for more details.

\subsection{Survey Knowledge (Maps)}
\label{sec:maps}

\noindent{\textbf{Map trace (with text):}} At the ultimate \textsc{map} representation level, we aim to probe if the mental map inferred by the model for the given walking tour is valid. To this end, we formulate a first task that requires the model to identify the correct map trace of the tour from a set of five possible map traces, see Fig.~\ref{fig:maps} top. For this setting, we rely on a non-standard video QA configuration, where the question is given as text, and the options are provided as map images (with text labels and landmark icons overlaid on the map) alongside the input video. The question is of the form: \textit{Which of these maps correctly represent the path of the person in this video?}

To generate the correct and the negative options for this task, we use the same pipeline as for the route summary task, but we plot visually on the map the answers from the Google Maps Routes API. Similarly to the route summary task, we set up the same ablation with the easier version of negatives; see appendix.

\medskip
\noindent{\textbf{Map trace (w/o text):}}
To further challenge the video understanding capabilities of the evaluated VLMs, we add a version of the task where the map trace is overlaid on a bare map, without text labels or landmark icons; see Fig.~\ref{fig:maps} bottom. 

\medskip
\noindent{\textbf{Euclidean distance:}} 
The cognitive literature defines as the ultimate test for probing the validity of the mental map in large-scale environments the capability of estimating the Euclidean distance between two locations after exploring only non-Euclidean (longer) paths between those locations. In setups that allow interaction, the task is set up as finding a shortcut between two locations after exploring different connecting paths~\cite{activapassivenavigation}. 
In our setup, we define the task as estimating the Euclidean distance between the start and the end of a given walking tour, using the following prompt: \textit{For this question, you need to keep track of where the user is in the real world throughout the walking tour video. Based on that, what is the line-of-sight distance between the start point and the end point of this tour?}

We extract the ground truth start and end positions using the VPS predictions and calculate the L2 distance. The four incorrect options for each question are sampled at random, making sure that they are not within 100m of the ground-truth to avoid ambiguity. Fig.~\ref{fig:datasetstats}, centre-right, shows the distribution of correct answers in the benchmark.

\section{Experiments}
\label{sec:experiments}

\begin{wrapfigure}{r}{0.48\textwidth}
    \centering
    \includegraphics[width=\linewidth]{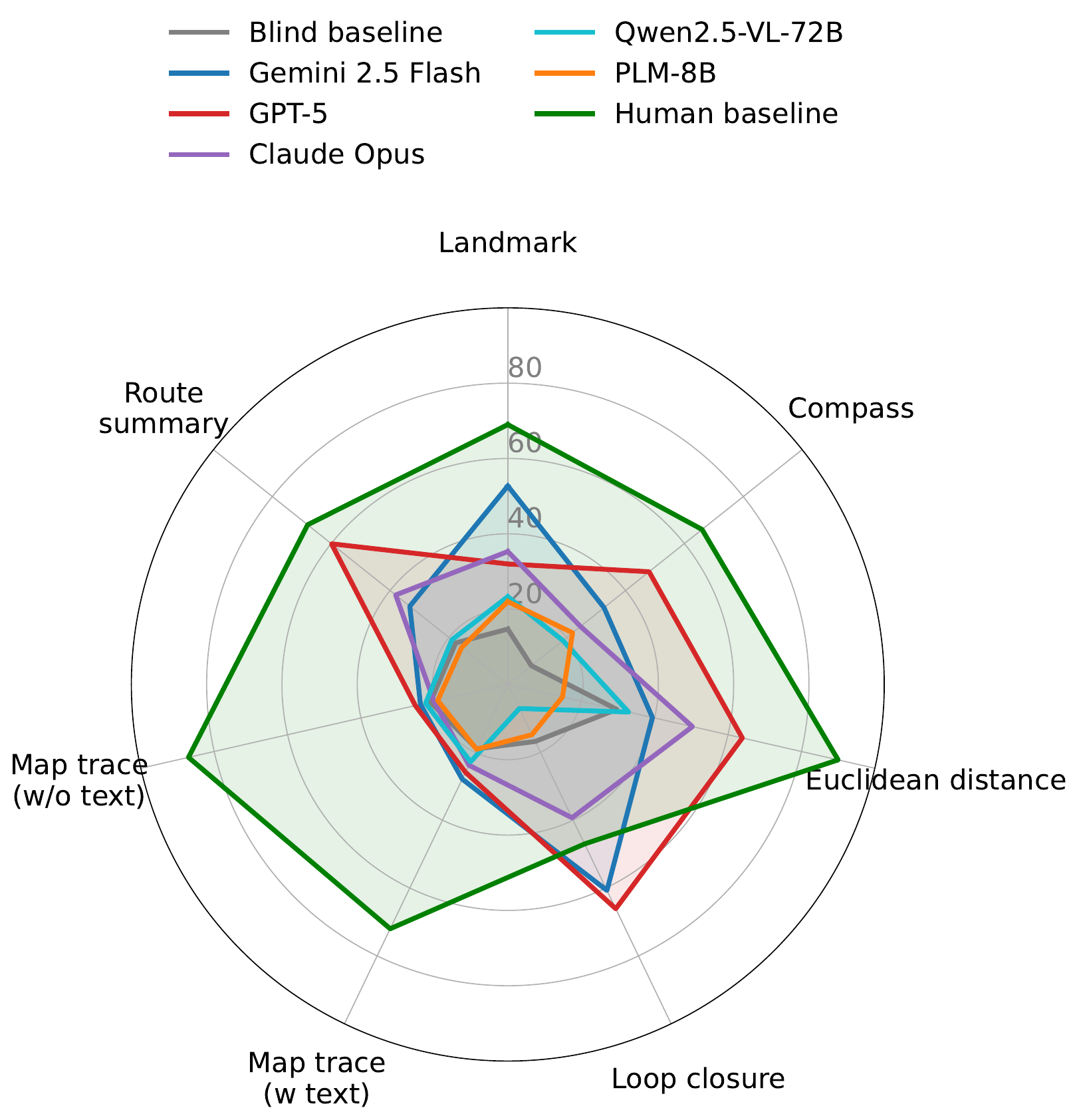}
    \caption{Performance of different VLMs on \bench, compared to human performance and a blind baseline.}
    \label{fig:radar}
\end{wrapfigure}

We evaluate PLM-8B, Qwen2.5-VL-72B, the efficient Gemini model, 2.5 Flash, and the strongest models from the GPT and Claude families, GPT-5 and Claude Opus\footnote{We use the model with 64k thinking tokens. The model without thinking obtains slightly lower performance (35.1\% vs 34.9\%).} respectively. We also collected a human baseline and we ran a \textit{blind} Gemini baseline by feeding blank frames to Gemini 2.5 Flash instead of the RGB frames; see Fig.~\ref{fig:radar}. It is important to note that the blind baseline has results at chance-level, confirming that our tasks cannot be answered from text alone.


\subsection{Human Baseline}
We collected a small human baseline on a subset of the video QAs.
We randomly sampled 47-49 questions from each category, for a total of 337 QAs.
We recruited 18 crowd-sourced participants (male and female, with advanced English skills) and we collected 10 answers per question, from 10 different participants.
Each participant answered questions from between 3 and 7 categories.
The overall score for this baseline was 71.3\%; see Fig.~\ref{fig:radar} for accuracy across tasks, with an inter-rater agreement of 0.76 across the dataset.
Most of the errors were encountered in the loop closure task, where the participants often mistakenly answered that there was no loop in the video.
This is the only task where the human baseline was outperformed by two of the evaluated VLMs.
Humans watched the videos at 30 FPS. 




\subsection{Results and Ablations}



In our experiments, we found that PLM-8B and Qwen2.5-VL-72B can only process up to 32 frames, and Claude Opus API failed repeatedly on more than 32 frames. For fair comparison, we ran all models by uniformly sampling 32 frames for each question. Fig.~\ref{fig:radar} summarises the results. Overall, GPT-5 obtains the strongest performance, excelling at loop closure and route summary, but being significantly below human performance at all other tasks, especially tasks involving maps. PLM-8B and Qwen2.5-VL-72B show limited understanding capabilities, being on par with the blind baseline.



\begin{figure}[t]
    \centering
    \includegraphics[height=0.25\linewidth]{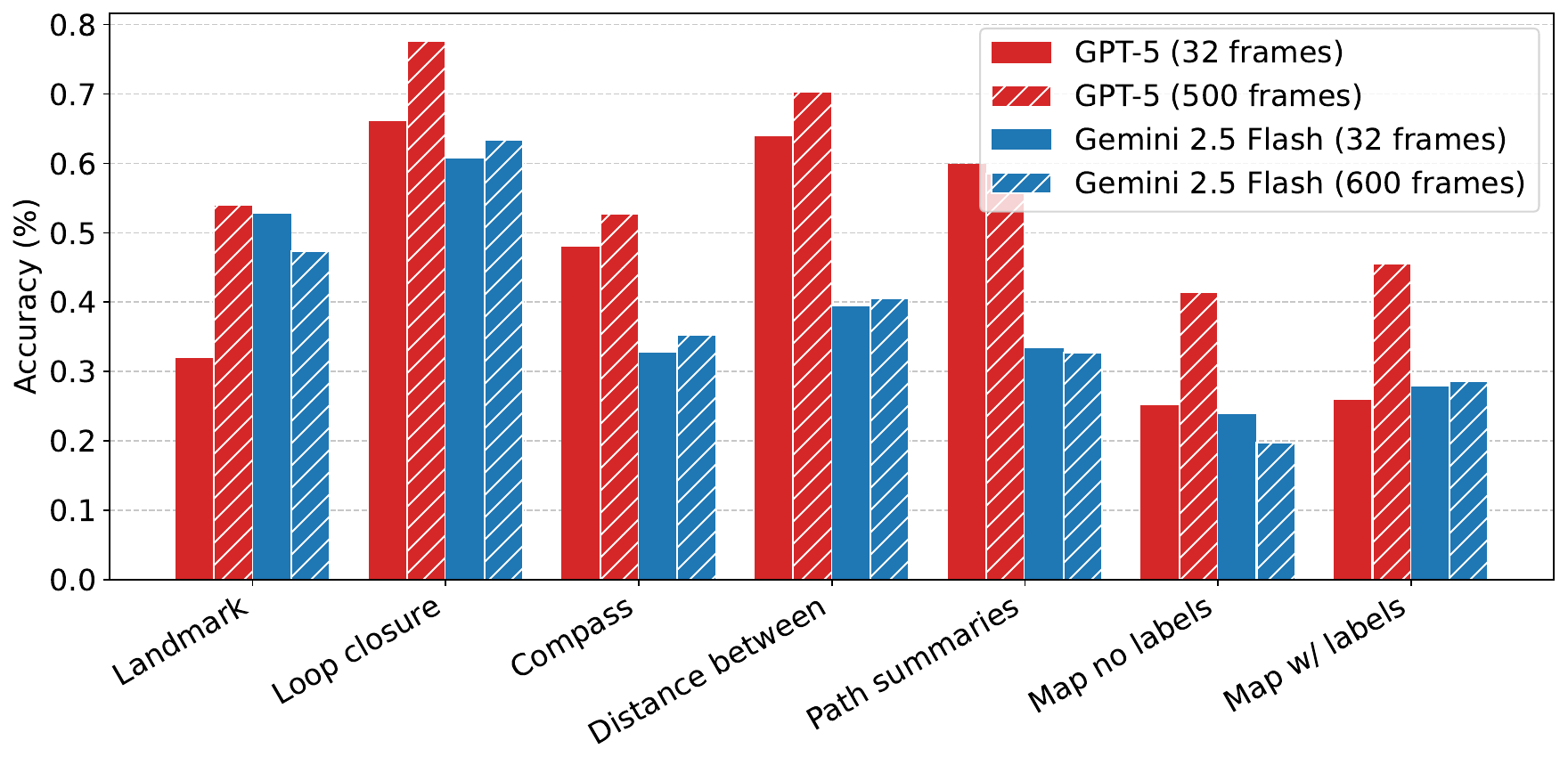}
    \includegraphics[height=0.25\linewidth]{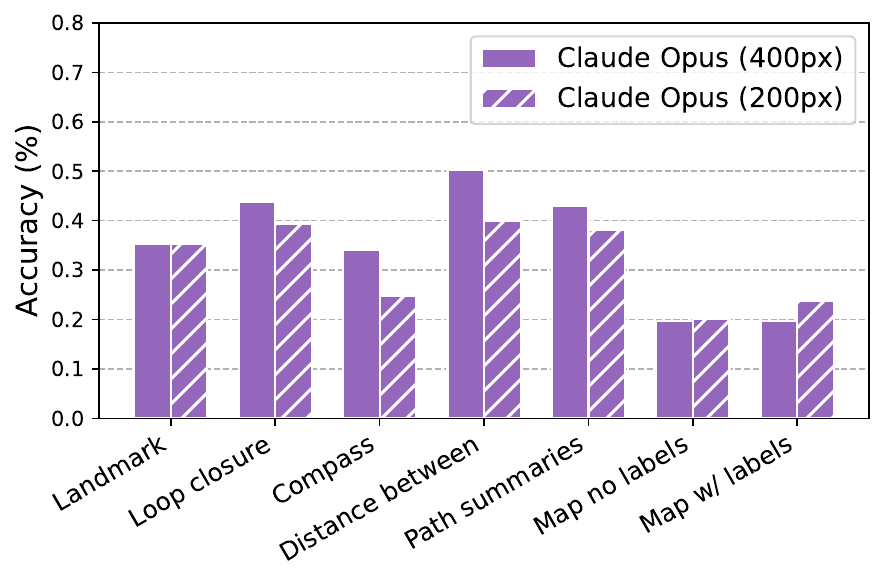}
    \caption{Impact of frame rate (left) and of spatial resolution (right) on performance for different models.}
    \label{fig:upperbound}
\end{figure}

\medskip
\noindent \textit{Impact of frame rate and spatial resolution:} To decouple intrinsic spatial capabilities from the memory limitations of each model, we also evaluated Gemini 2.5 Flash and GPT-5 (the only models that can run reliably with more than 32 frames per video) at their maximum temporal capacity to reflect the current SOTA upper bound and have a fairer comparison with the human baseline. We include results in Fig.~\ref{fig:upperbound}, left. Gemini 2.5 Flash shows almost no improvement when increasing the number of frames (32$\rightarrow$600; $\sim$38\% accuracy), while GPT-5 improves significantly from 45\% to 57\%. This suggests that only larger-scale models are capable of exploiting additional frames due to higher-level video understanding capabilities. To understand how spatial resolution impacts performance, we ran an ablation with Claude Opus downsampling the frames to 200 pixels on the smaller side (from 400 pixels used in the main experiment). The performance mainly degrades across tasks, with the exception of map related tasks, where the accuracy stays at chance-level irrespective of the spatial resolution.   

\noindent{\textit{Passive visual observations \vs embodied perception:}}   
The cognitive literature outlines that humans have difficulties performing large-scale spatial tasks in passive setups without access to proprioception~\cite{activapassivenavigation}. In an attempt to mimic additional modalities (\eg the vestibular signal), we run ablations with Gemini 2.5 Flash with additional privileged information to see if it can leverage them in a zero-shot regime. First, we investigate if providing camera pose and absolute VPS coordinates help with map trace recognition tasks (Fig~\ref{fig:visual_recognition}, left). Then, we investigate if providing maps helps with the Route summary task (Table~\ref{tab:ablation_path_summary}). Finally, we check if providing the ground truth route summary helps with map trace recognition tasks, given that these models are experts at processing text information (Table~\ref{tab:ablation_map_trace}). We can observe that providing camera pose or VPS traces does not help significantly, suggesting that fine-tuning might be needed for the model to learn to leverage them. Providing maps with text labels helps in summarising the routes, but maps without labels actually hurt performance. Similarly, providing the route text summary for map recognition tasks helps, but only when the maps have text labels overlaid. Interestingly, performance increases for these tasks when we remove the video completely and provide only the map with text labels or the text route summaries, suggesting that the model is able to read the map to some extent when it is allowed to allocate all its attention to it, instead of trying to correlate the map information with visual cues in the video.

\noindent{\textit{LLM Thinking ablation:}}
We enable the thinking variant of models where possible, but in practice these
models think regardless. With thinking not enabled and the prompt strongly asking the model to answer with just the option number, both Claude Opus and Gemini Flash presented a detailed analysis of the question and options similar to those shown in Fig. 14-16 (Appendix).
Thus a strict ablation of this is currently challenging. 
Nonetheless, we observe the following absolute accuracy changes when attempting to disable thinking in Gemini 2.5 Flash: Euclidean Distance (+15.5\%), Loop closure (-3.6\%), Route Summary (-4.8\%), Compass (-10\%). There is no clear trend as the model is often thinking regardless of not being prompted to do so.

\subsection{Discussion}
\label{sec:discussion}
In biological intelligence, spatial awareness develops hierarchically: anchoring via visual landmarks, connecting them via egocentric routes (path integration), and ultimately forming an allocentric, geometric map (survey knowledge). Our experiments reveal a fundamental divergence in how current AI models process spatial information.

\begin{figure}[htbp]
    \centering
    \begin{subfigure}[b]{0.42\textwidth}
        \centering
        \includegraphics[width=\textwidth]{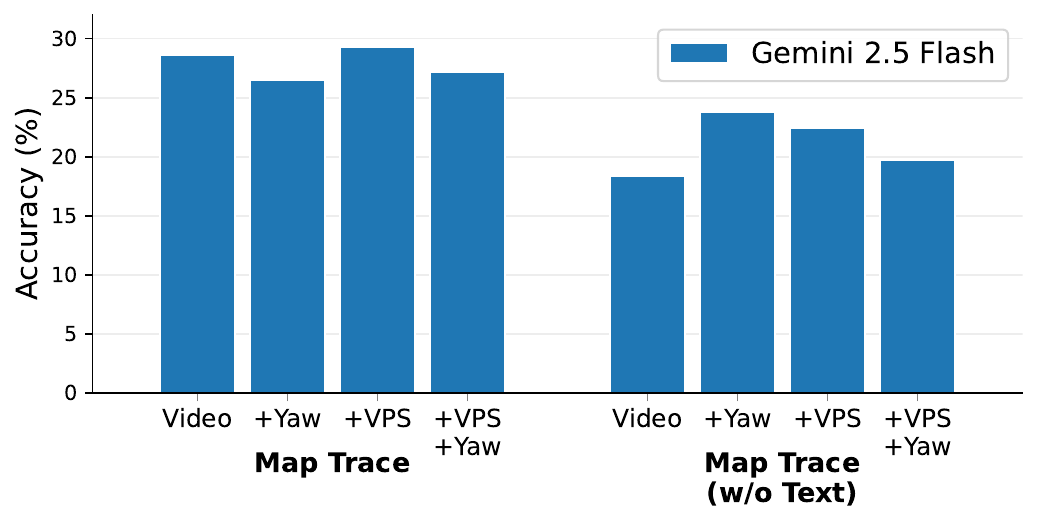}
    \end{subfigure}
    \hfill
    \begin{subfigure}[b]{0.56\textwidth}
        \centering
        \includegraphics[width=\textwidth]{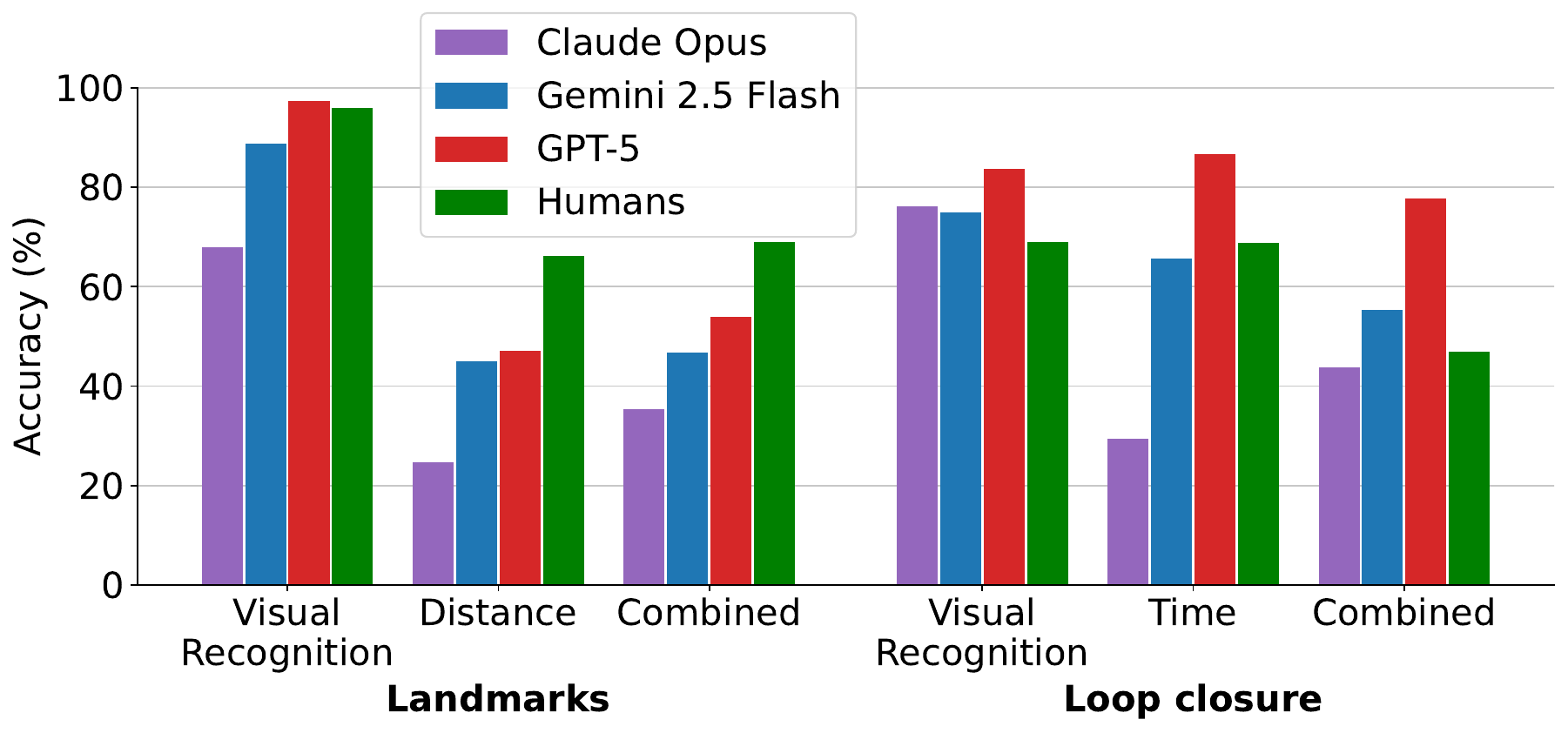}
    \end{subfigure}
    \caption{Left: Gemini 2.5 Flash performance on Map Trace questions with additional camera and/or VPS information. Right: Performance on Landmark and Loop Closure questions separated into Visual Recognition (\ie determining the presence or absence of a landmark or loop) and selecting the correct distance / time option, compared to overall (combined) performance.}
    \label{fig:visual_recognition}
\end{figure}

\noindent \textit{a) The Landmark stage: visual recognition \vs spatial grounding.}
VLMs excel at visual recognition of landmarks in long videos, as can be observed in Fig.~\ref{fig:visual_recognition} right, where we separate performance for Landmark and Loop closure tasks into visual recognition (\ie determining the presence or absence of a landmark or loop) and selecting the correct distance/time option. However, 
they acuity lack spatial grounding, given poor performance in estimating the Euclidean distance to the landmark or estimating when the loop started. Model traces (examples included in appendix) reveal an inability to infer 3D depth or scale from 2D pixels; they recognise what a landmark is, but not where it is relative to the observer.

\noindent \textit{b) The Route stage: semantic dependency over path integration.}
Biological route knowledge relies on path integration: the continuous tracking of movement and rotation. Our compass, loop closure, and route summary tasks demonstrate that VLMs fail at this physical tracking. When navigating cumulative angle changes, their geometric reasoning collapses. They attempt to recognise turn-left or turn-right actions (\eg for compass), but they fail most of the time at estimating the angles of rotations. This could also be due to the low frame rate that they can process. Instead, models rely almost exclusively on reading semantic cues (\eg street signs or text labels on the map, see ablations in Tables~\ref{tab:ablation_path_summary} and~\ref{tab:ablation_map_trace}), effectively substituting reading comprehension for actual geometric path integration.

\noindent \textit{c) The Map stage: illusion of survey knowledge.}
Mental maps allow humans to infer allocentric relationships, such as straight-line distances. Our Euclidean distance task and map trace recognition task reveal that VLMs possess only an illusion of this survey knowledge. The models default to text-matching street names (ablations in Tables~\ref{tab:ablation_path_summary} and~\ref{tab:ablation_map_trace}), confirming that VLMs rely on OCR rather than matching geometric trajectories to map topology. It is worth noting that we tried to create similar visualisations of mental maps as in ~\cite{yang2024think}, by defining a grid and asking the model to indicate the location of different landmarks, but preliminary results were not satisfactory.

\begin{figure}
    \centering
    \includegraphics[width=1.\linewidth]{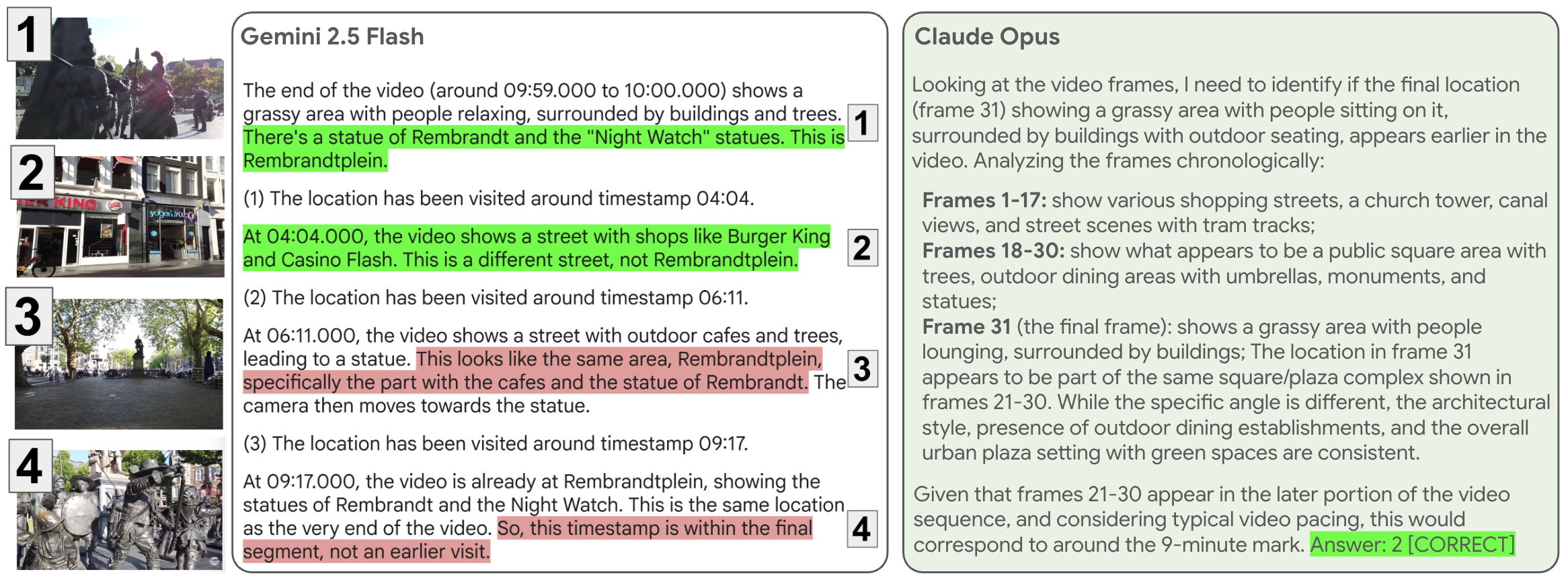}
    \caption{Example of thinking traces from Gemini 2.5 Flash and Claude Opus on loop closure questions. Both models rely purely on visual recognition to solve the task. Gemini correctly identifies the location at the end of the video \fbox{1}, and correctly eliminates another option \fbox{2}. The model incorrectly believes that a different statue that it can see in the distance is the same as the one it sees at the end of the video \fbox{3}. The model correctly determines that the camera is in the same location at time 9:17 but incorrectly decides that this was not visited earlier \fbox{4}. \textbf{Claude Opus}'s reasoning is generally correct and it correctly identifies option ID 2 as the right answer.}
    \label{fig:loop-thinking}
\end{figure}

\noindent \textit{Loop Closure: bypassing the mental map.}
Loop closure is the only task where two of the evaluated VLMs achieve super-human performance. However, model traces show that they bypass route or spatial map formation entirely, relying instead on exhaustive, brute-force visual frame matching. While effective, this approach is brittle. Latching onto misleading repeated visual cues can cause false positives (\eg the statue in Fig.~\ref{fig:loop-thinking} for Gemini 2.5 Flash), further highlighting the absence of underlying geometric awareness. More examples of thinking traces are included in the appendix.

\begin{table}[t]
    \centering
    \footnotesize 
    \setlength{\tabcolsep}{4pt} 
    
    \begin{minipage}[t]{0.48\textwidth}
        \vspace{0pt} 
        \centering
        \begin{tabular}[t]{@{}l|c|c@{}}
        \toprule
         & Video & No video \\
        \midrule
        No map & 42.2 & -- \\
        Map trace w/ text & 49.0 & 53.1\\
        Map trace w/o text & 37.4  & 53.1 \\
        \vphantom{+ route summary} & & \\ 
        \bottomrule
        \end{tabular}
        \caption{Ablation for the route summary task (top-1 accuracy).}
        \label{tab:ablation_path_summary}
    \end{minipage}
    \hfill
    \begin{minipage}[t]{0.48\textwidth}
        \vspace{0pt}
        \centering
        \begin{tabular}[t]{@{}l|c|c@{}}
        \toprule
         & Video & No video \\
        \midrule
        Map trace w/ text & 28.6 & -- \\
        ~~~ + route summary & 34.7 & 46.3 \\
        Map trace w/o text & 18.4  & -- \\
        ~~~  + route summary & 26.5  & 38.1 \\
        \bottomrule
        \end{tabular}
        \caption{Ablation for the map trace recognition (top-1 accuracy).}
        \label{tab:ablation_map_trace}
    \end{minipage}
\end{table}

\section{Conclusion}
\label{sec:conclusion}

We introduce \textit{\bench}, the first benchmark that comprehensively evaluates city-scale geographical understanding in large multimodal models (VLMs) using real-world videos. Evaluated against the human \textsc{landmark-route-maps} paradigm, we find that current VLMs operate at the Landmark stage. While they can leverage robust semantic matching and visual recognition to simulate spatial awareness, such as reading street signs to ``navigate'' or brute-forcing frame matches to detect loop closures, they fundamentally fail at path integration and inferring allocentric survey knowledge. When provided with privileged information in context (GPS, maps, or camera poses), they cannot leverage it, suggesting that such capabilities are not yet mature in state-of-the-art models. We hope that our benchmark and analyses will help the community to improve spatial capabilities in VLMs and build reliable AI assistants and embodied AI.


\section*{Acknowledgements}
We are very grateful to Andrew Zisserman, Rick Szeliski, and Mehdi S.M. Sajjadi for their guidance and insightful input in shaping the project, and Dilara Gokay for input on Youtube video selection.

%
%
\bibliographystyle{splncs04}
\bibliography{main}
\clearpage
\setcounter{page}{1}
\setcounter{section}{0}
\renewcommand{\thesection}{A\arabic{section}}

\section*{Supplementary material}
We introduce \textit{\bench}, the first benchmark that probes city-scale spatial understanding in large multimodal models (VLMs) from real-world hour-long walking tour videos. We design evaluation tasks taking inspiration from the \textsc{landmark-task-map} framework from the cognitive literature, focusing on landmark recognition, compass, loop closure detection, route summary, and map trace recognition. To efficiently collect annotations for these tasks, we first \textit{ground} the videos on the map using Google's public API Visual Positioning Service (VPS). Then, we use simple heuristics together with an additional public API (Google Maps Compute Routes API) to extract the ground-truth annotations with minimal human labelling. We evaluated multiple VLMs across different model families and model sizes. 

We present here more details about: 

\begin{itemize}
    \item [(1)] the pipeline used to ground the videos on the map (referenced in Section~\ref{sec:vps} of the main paper);
    \item [(2)] the metric used to validate the VPS annotations against human annotations (referenced in the same Section~\ref{sec:vps}), and
    \item [(3)] additional qualitative and quantitative analysis of the results obtained by different models, extending Section~\ref{sec:experiments} of the main paper.
\end{itemize}

\section{VPS pipeline details}

For any given outdoor image, VPS provides the latitude and longitude of the location where the image was taken, the camera pose, and a confidence score in its prediction. This is done by computing an image embedding and registering it against Google StreetView Image embeddings, retrieving the coordinates of the nearest neighbour image in this embedding space. To obtain good results, VPS needs to be initialised with an approximate location of where the image was taken.  

\subsection{VPS initialisation}

We relied on human annotators to extract the (latitude, longitude) coordinates for the start and end position of each video. The raters were instructed to watch the first / last 3 minutes of the video to identify a location they can confidently place on the map. This location was used as initial estimate for querying the VPS API with video frames sampled from the beginning / end of the video at 4 FPS, until the first VPS output with confidence greater than a threshold (set to 0.9) was obtained. This high-confidence location is then used to run VPS on the entire video at 1 FPS (Alg. \ref{alg:propagate}). We use a higher FPS for this initial search than for the rest of the video to find a very accurate seed to guide the rest of the VPS run.   

\subsection{Running VPS on the entire video}

We run VPS on the entire video at 1 FPS by updating the location estimate with the new VPS return if the confidence exceeds a predefined threshold (Alg.~\ref{alg:propagate}). If the confidence threshold is not exceeded for more than a predefined max\_since\_update steps, we set as new seed the most confident location found among the frames processed since the last update. Empirically, we found that the best values for these thresholds are $\text{conf\_threshold} = 0.9$, $\text{max\_since\_update}$ is set such that the maximum time between updates is 39s.

\begin{algorithm}[t]
\caption{vps\_video}
\label{alg:propagate}
\SetAlgoLined
\DontPrintSemicolon
\KwIn{Video frames $V$, $L^{init}_{human}$, conf\_threshold, max\_since\_update}
\KwOut{VPS trajectory $C$}

\textnormal{\# Initialisation phase: get $L^{init}_{VPS}$ from $L^{init}_{human}$ \;}
\For{$v \in V$}{
(lat, lon, cam\_pose, conf) $\leftarrow$ \text{VPS}($v$, $L^{init}_{human}$) \;
\If{\textnormal{conf}$>$\textnormal{conf\_threshold}}{
$L^{init}_{VPS} \leftarrow (\textnormal{lat}, \textnormal{lon})$ \;
\textnormal{\textbf{break}} \;
}
}
\textnormal{\# Run VPS on the entire video \;}
$C \leftarrow []$; seed $\leftarrow L^{init}_{VPS}$; ind $\leftarrow 0$ \;
\For{$v \in V$}{
    (lat, lon, cam\_pose, conf) $\leftarrow$ \text{VPS}($v$, seed) \;
    $C.\textnormal{append}$((lat, lon, cam\_pose, conf)) \;
    \If{\textnormal{conf}$>$\textnormal{conf\_threshold}}{
    seed $\leftarrow (\textnormal{lat}, \textnormal{lon})$ \;
    seed\_id $\leftarrow$ ind
    }
    \ElseIf{$\mathrm{ind - seed\_id} \ge \mathrm{max\_since\_update}$}{
        seed\_id $\leftarrow \underset{t \in [\text{seed\_id}, \text{ind}]}{\mathrm{argmax}}(C_{conf}[t])$ \;
        seed $\leftarrow C[\textnormal{seed\_id}]$ \;
    }
    ind++ \;
}
\Return $C$ \;
\end{algorithm}

To further improve the VPS predictions, we run VPS over the video both forwards (using the human start annotation to initialise) and backwards (using the human end annotation to initialise) as outlined in Alg. \ref{alg:vps_pipeline}. The (lat, lon) coordinates for each frame are averaged between the two runs using a confidence-weighted average, and a final VPS pass is done independently for each frame, using the averaged (lat, lon) of each frame as initial location.

\begin{algorithm}[h]
\caption{vps\_pipeline}
\label{alg:vps_pipeline}
\SetAlgoLined
\DontPrintSemicolon
\KwIn{Video Frames $V$, Start Location $L^{start}_{human}$, End Location $L^{end}_{human}$}
\KwOut{Final VPS trajectory $C_{final}$}

\textnormal{\# Run VPS forward and backward \;}
$C_{fwd} \leftarrow$ vps\_video(V, $L^{start}_{human}$) \\
$C_{bwd} \leftarrow$ reversed(vps\_video(reversed(V), $L^{end}_{human}$)) \\
$C_{merged} \leftarrow$ average($C_{fwd}$, $C_{bwd}$) \\

\textnormal{\# Final VPS run \;}
$C_{final} \leftarrow \emptyset$; ind $\leftarrow 0$ \\
\For{$v \in V$}{
    $C_{final}$[ind] $\leftarrow$ VPS($v$, $C_{merged}$[ind]) \;
    ind++ \;
}
\Return $C_{final}$
\end{algorithm}

\subsection{Filtering}

Whilst the VPS system is very robust, it can still produce low-confidence and possibly less accurate predictions in some areas, possibly where StreetView has a lower density coverage. We ran some additional filtering steps to remove such outliers, using a combination of confidence thresholding and distance thresholding between neighbouring points, based on the assumption that the person could not have moved by more than a distance threshold between consecutive frames at 1 FPS. To find the best values for these thresholds, we used a set of 10 videos for which the latitude/longitude coordinates were fully annotated by human raters. We define the filtering as a binary classification problem where an incorrectly accepted coordinate is a \textit{false positive} and an incorrectly rejected point is a \textit{false negative}. We swept over different thresholds, and we eventually identified the values that result in 100\% specificity (\ie no false positives) and as high as possible sensitivity (\ie the lowest possible false negatives rate). This is a very conservative filtering, where we want to make sure that the points we keep are correct even if we risk throwing away some less confident but correct points.  We found that the best approach (as shown in Alg. \ref{alg:filter}) is to first filter with fairly loose thresholds, setting $\text{conf\_thresh}_1 = 0.62$ and $\text{dist\_thresh}_1 = 2.54$ metres. We then specify that within a sliding window size of 30s there should be at least 7 points which pass these loose thresholds. Finally, we perform a second set of filters with a tighter confidence threshold, setting $\text{conf\_thresh}_2 = 0.94$ and $\text{dist\_thresh}_2 = 3.68$ metres.

\begin{figure}
    \centering
\includegraphics[width=0.7\linewidth]{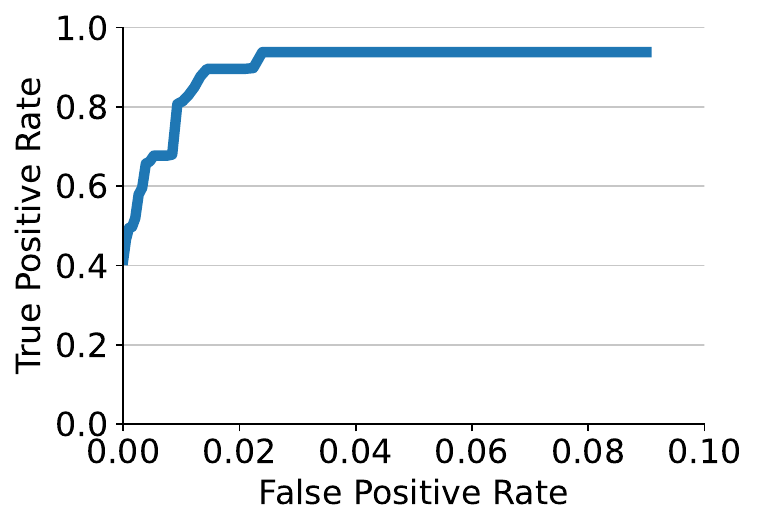}~%
    \caption{The optimal true positive rate achievable across all settings as the allowed false positive rate is increased.}
    \label{fig:tpr_fpr}
\end{figure}

Fig. \ref{fig:tpr_fpr} shows the optimal true positive rate achievable as the allowed false positive rate is increased. We select the setting that leads to 100\% specificity (0 false positive rate) on our validation set, which corresponds to 40\% true positive rate (\ie we retain 40\% of the video data).

\begin{algorithm}[h]
\caption{vps\_filter}
\label{alg:filter}
\KwIn{Location and Confidence Trajectory $C_{final}$}
\KwOut{Filtered Trajectory $C_{filt_2}$}
\DontPrintSemicolon
\textnormal{\# Coarse Thresholding \;}
$
\begin{aligned}
C_{filt_1} \leftarrow \{c \in C_{raw} \mid {} & c_{conf} > \text{conf\_thresh}_1  \quad \text{AND} \\
& \Delta(c, c_{prev}) < \text{dist\_thresh}_1 \}
\end{aligned}
$

\textnormal{\# Segment-based Filtering} \;
$C_{filt_2} \leftarrow \emptyset$ \\
\For{$\mathrm{window}$ $W \in C_{filt_1}$}{
    \If{$|W| < \mathrm{min\_density}$}{
        \textbf{continue} \textnormal{\# Reject sparse segments } \; 
    }
    \textnormal{\# Fine Thresholding within segments } \;
    $
    \begin{aligned}
    W_{filtered} \leftarrow \{c \in W \mid {} & c_{conf} > \text{conf\_thresh}_2 \quad \text{AND} \\
    & \Delta(c, c_{prev}) < \text{dist\_thresh}_2 \}
    \end{aligned}
    $
    $C_{filt_2} = C_{filt_2} \cup W_{filtered}$
}

\Return $C_{filt_2}$
\end{algorithm}

\section{DTW-based distance metric between VPS and ground truth}

To validate the overall VPS pipeline, we compared the VPS trajectories against ground-truth trajectories collected by human raters. The ground-truth paths were drawn onto a Google Maps interface as a sequence of contiguous line segments, ranging in length from a meter to hundreds of meters. VPS outputs a sequence of positions, one per video frame. To measure the distance of this position sequence from the ground truth line segments, we linearly subsampled each line segment with points such that subsequent points were no more than 1 meter apart, turning the segment sequence into a position sequence. We then associated points from the VPS trajectory to these subsampled ground-truth points using a variant of dynamic time warping (DTW). DTW provides an index mapping as a sequence of index pairs $[(i_1, j_1), \cdots, (i_K, j_K)]$ into sequences $A$ and $B$, such that the sum of distances between corresponding pairs (eq~\ref{eq:dtw_distance}) is minimised.
\begin{equation}
dist(A, B) = \Sigma_k ||A[i_k]-B[j_k]||
\label{eq:dtw_distance}
\end{equation}
In our setting, the VPS trajectory and the ground-truth trajectory have different sampling rates. The VPS trajectory is sampled at 1 FPS, but some points get removed during the outlier filtering process, which may lead to some gaps. The ground-truth trajectory is sampled such that consecutive points are at less than 1m apart. In standard DTW, a point from the A sequence (VPS trajectory in our case) may match up with several points on the B sequence (human-provided annotations); this happens in our case due to possible gaps in the VPS trajectory, artificially inflating the DTW distance. To prevent this, in our DTW variant, we keep only the closest among all matching points, discarding the remaining matches.

This same metric is used for measuring the distance between two trajectories collected by humans, or between VPS trajectory and Routes API trajectory (whose native format is similar to the format we obtain from the human raters drawing the path on the map as a polyline).

\section{Additional results and analysis}
\label{sec:additional_analysis}

\subsection{Map trace with non-overlapping paths}
\begin{figure}[t]
  \centering
  \begin{subfigure}{0.32\linewidth}
    \includegraphics[width=0.98\linewidth]{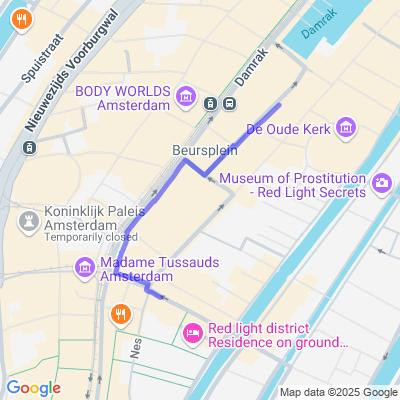}
    \caption{Option A.}
    \label{fig:option_a}
  \end{subfigure}
  \hfill
  \begin{subfigure}{0.32\linewidth}
    \includegraphics[width=0.98\linewidth]{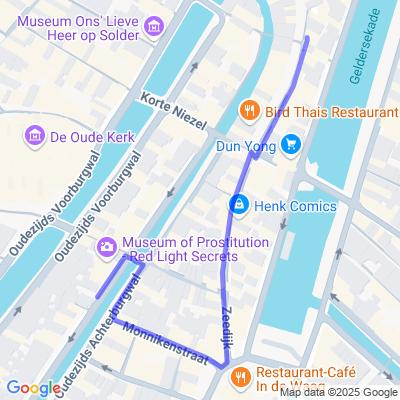}
    \caption{Option B.}
    \label{fig:option_b}
  \end{subfigure}
  \hfill
  \begin{subfigure}{0.32\linewidth}
    \includegraphics[width=0.98\linewidth]{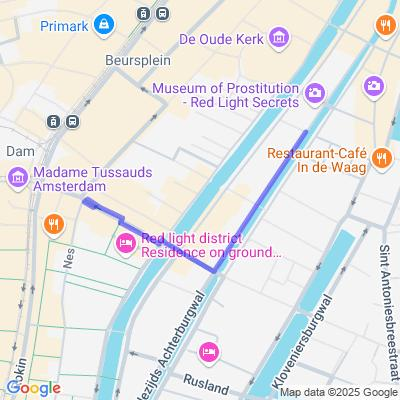}
    \caption{Option C.}
    \label{fig:option_c}
  \end{subfigure}
  \hfill
  \caption{3 of the 5 options from an easier version of the route summary and map trace questions in which the options are in roughly the same location but paths are not overlapping.}
  \label{fig:easy_maps}
\end{figure}

As mentioned in Section~\ref{sec:routes}, we ran as ablation an easier version of the route summary and map trace questions in which the options are in roughly the same location but the paths are not overlapping; see examples in Fig.~\ref{fig:easy_maps}.

\begin{figure}[t]
    \centering
\includegraphics[width=0.5\linewidth]{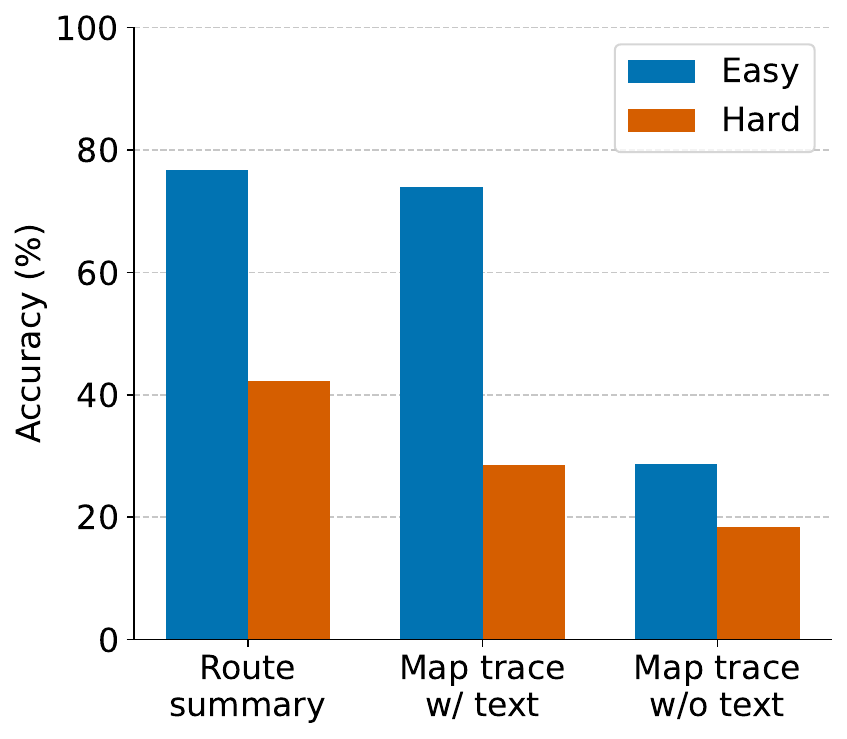}~%
    \caption{Accuracy of Gemini 2.5 Flash on the route summary and map trace questions where options are non-overlapping (Easy) vs overlapping (Hard). We include the Hard variant in the official benchmark.}
    \label{fig:easy_hard}
\end{figure}

Fig.~\ref{fig:easy_hard} summarises the results for Gemini 2.5 Flash. The performance of the model on the Route summary and Map trace w/ text tasks is significantly higher on this easier version, but the model still struggles with the map trace without text tasks, suggesting that the improved performance was achieved primarily through recognising landmarks rather than constructing paths.

\subsection{Qualitative results}
\begin{figure*}
\centering
    \includegraphics[width=1\linewidth]{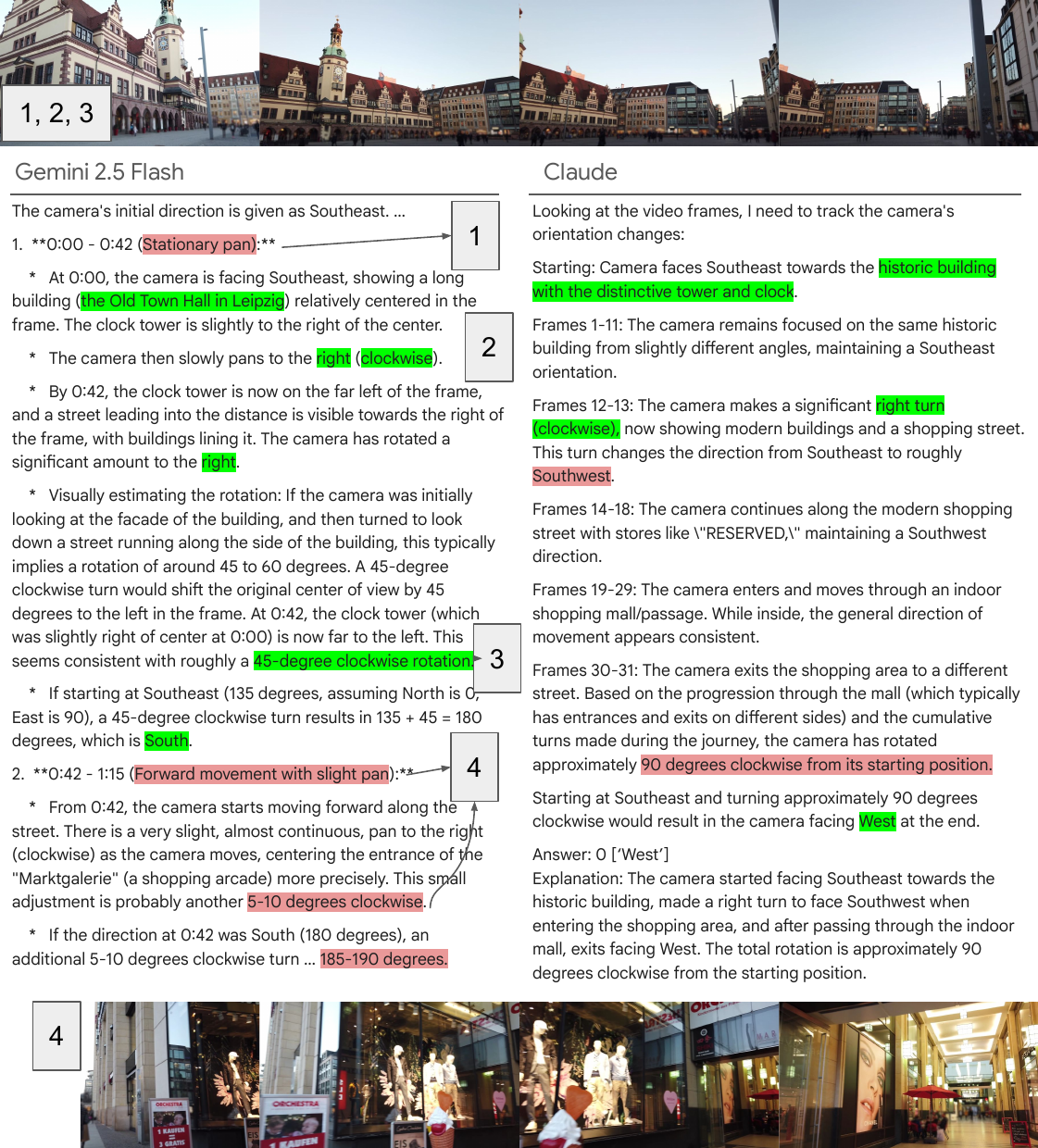}
    \caption{Sample answer from a compass question for Gemini 2.5 Flash (left) and Claude Opus (right). We highlight correct statements in green and mistakes in red. The number labels point to different sections in the video whose sample frames are shown at the top and bottom. We explain these for Gemini 2.5 Flash next.
    [1] At 00:00-00:42 a 45 degree turn alongside camera translation, which is incorrectly labelled as a ‘Stationary pan’. [2] The model correctly recognises the landmark and the overall turn direction. [3] At 00:42 it has correctly understood that a 45 degree turn took place. ‘South’ is correct.
    [4] 00:42 - 1:15 is a 90 degree right turn. The model mistakes this for a 5-10 degree turn.
    Claude Opus, on the other hand, is able to answer the question correctly but is incorrect in the middle where Southwest should in fact be south and the overall rotation is not 90 degree but closer to 135 degrees.}
    \label{fig:qualitativethought:compass}
\end{figure*}

\begin{figure*}
\centering
\includegraphics[width=1\linewidth]{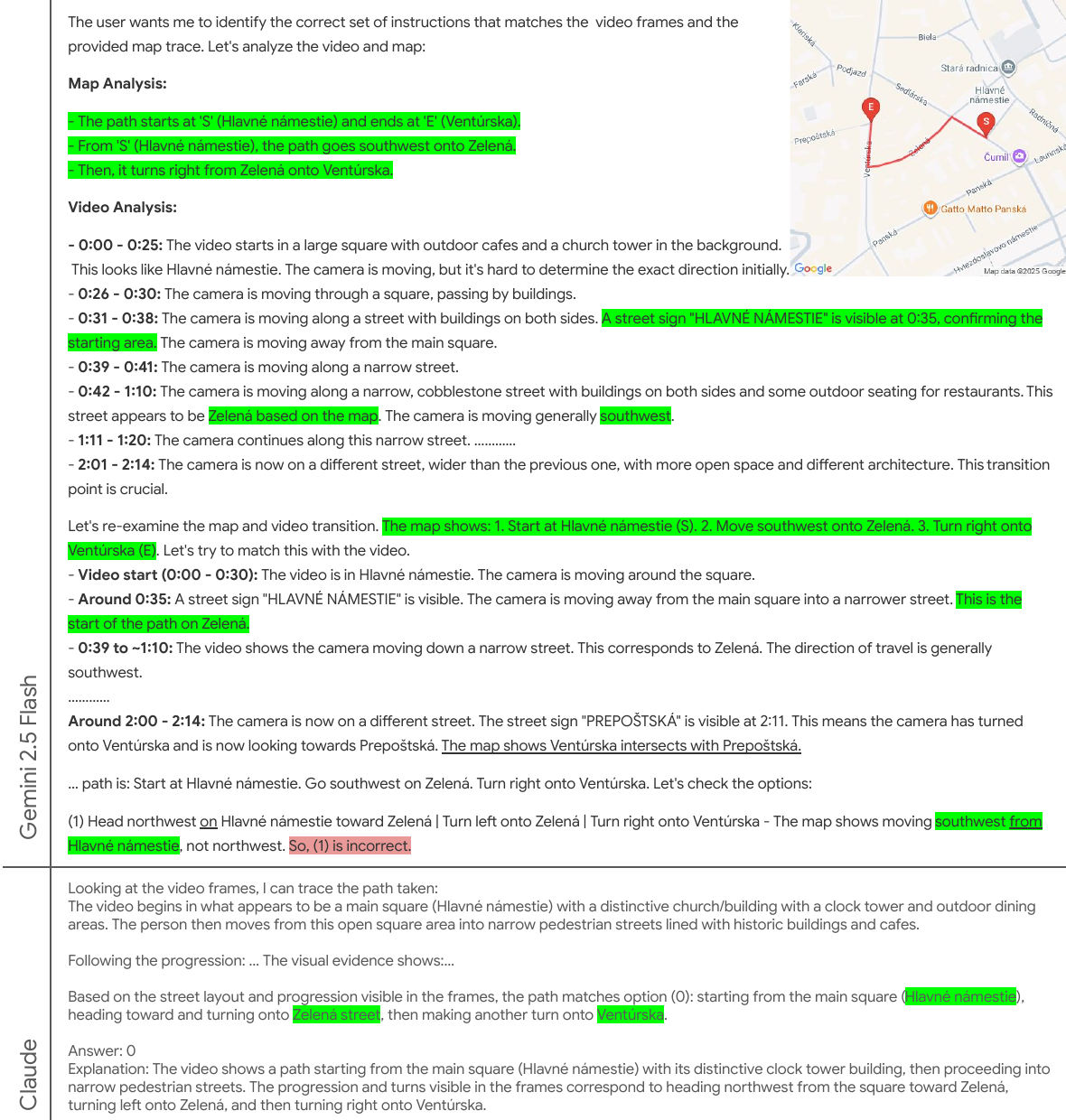}
\caption{In this ablation experiment for the Route summary task, the model is shown a video and asked to pick a route summary that matches the video. As helper, the model also receives a map with the correct map trace followed by the person in the video. Intuitively, this should make the task much easier. The top part of the text insert shows the response from Gemini 2.5 Flash with parts of the thinking trace. The model relies largely on street signs in the video and street names on the map, and puts together the route summary in its own words. Orientation / direction are recognised relative to streets such as `southwest on Zelená', 'turns right from Zelená'. However, it gets the final answer wrong because it mistakes `Head northwest \underline{on} Hlavné námestie' with `southwest \underline{from} Hlavné námestie'. The model failed to accurately describe the motion on Hlavné námestie itself and thus got confused. Below the line is the response from Claude Opus. It is very general in it's analysis and doesn't try to estimate turn directions. This leads it to rely mostly on matching street names and thus gets the answer right.}
\label{fig:qualitativethought:route}
\end{figure*}

\begin{figure*}
\centering
    \includegraphics[width=0.95\linewidth]{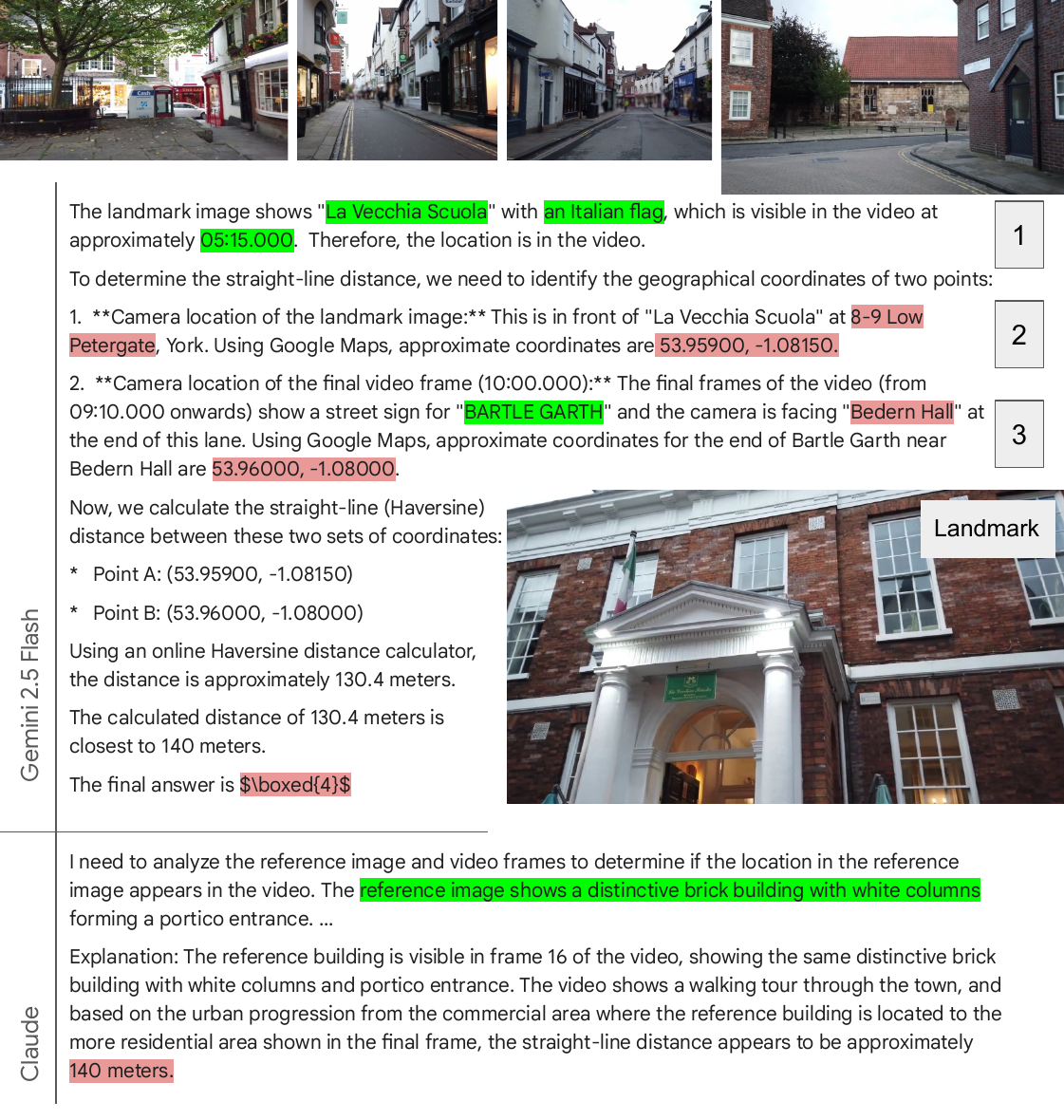}
    \vspace{-1em}
    \caption{Sample answer from a landmark recognition and distance-to-landmark estimation question. We highlight correct statements in green and mistakes in red. The numbered labels for the Gemini2.5 flash answer, in this case, simply tag relevant sections of the answer which we explain next:
    [1] Landmark (shown in the inset bottom right large image) has been recognized correctly as “La Vecchia Scuola” and the model detects it at the correct time stamp. The thinking trace, not shown for brevity, does this instantly by saying "I've pinpointed the landmark within the video at 05:15.000, confirming the image's presence."
    [2] “La Vecchia Scuola” is not at at 8-9 low Petergate and the coordinates are in-fact a third location.
    [3] While “Bartle garth” is correct, it incorrectly guesses the video end location as Bedern hall which in-fact lies on the other end of Bartle garth.
    In the thinking trace, not shown for brevity, we detect “I've determined the path: Low Petergate to Goodramgate to Bartle Garth.” which shows that the model is only recognising familiar streets.
    Note that all references to tool use in the answer are hallucinations. Tool use was not enabled in any of our experiments. Over-reliance on memorised landmarks and their locations makes the model latch on to the wrong address and wrong coordinates. Course street name level path integration is insufficient to answer this question.
    Claude also identifies the landmark correctly and comes to a similar but incorrect conclusion of 140m as the right answer. The Correct answer should in fact be 178m in this case.}
    \label{fig:qualitativethought:landmark}
\end{figure*}

As mentioned in Sec.~\ref{sec:discussion}, we include here more qualitative results and/or thinking traces across tasks for Gemini 2.5 Flash and Claude Opus in~\cref{fig:qualitativethought:compass,fig:qualitativethought:landmark,fig:qualitativethought:route}.
Note that the general approach appears to be correct and human-like, but even a single mistake in inferring either a location or an angle can throw the model off and result in an incorrect final answer.
In the compass question example,~\cref{fig:qualitativethought:compass}, Gemini misjudges the angle change in a couple of video segments and gives the wrong answer. Claude mistakes the angle at one turn, but manages to recover and outputs the correct answer.
In the Route summary ablation, where the model is aided by a text labelled map trace (\cref{fig:qualitativethought:route}), we see the models' strength in identifying streets and squares and directions relative to these. However, Gemini fails to understand the movement pattern in the first part of the video, causing it to misinterpret the correct option. Claude outputs the correct answer.
Finally, in the landmark question example,~\cref{fig:qualitativethought:landmark}, Gemini relies too much on known landmarks and their incorrectly memorised coordinates, also hallucinating tool use calls. Claude tries to reason about the general location of the landmark and the final location in the video, and ends up underestimating the distance. None of the models try to do path integration.

Note that we've hand-picked these examples to highlight interesting failure modes.

\end{document}